\documentclass[journal]{IEEEtran}
\usepackage{cite}

\ifCLASSINFOpdf
\usepackage[pdftex]{graphicx}
\usepackage{amssymb}
\else
\fi

\usepackage{amsmath, amssymb}
\usepackage{makecell}
\usepackage{multirow}

\begin{document}

%
\title{Video Semantic Segmentation with \\ Distortion-Aware Feature Correction}
%
%
%

\author{Jiafan Zhuang, Zilei Wang,~\IEEEmembership{Member,~IEEE}, Bingke Wang 
	\thanks{This work was supported in part by the National Natural
		Science Foundation of China under Grant 61673362 and Grant 61836008,
		in part by Youth Innovation Promotion Association CAS (2017496), and in part by
		the Fundamental Research Funds for the Central Universities.}
	\thanks{J. Zhuang, Z. Wang, and B. Wang are with National Engineering Laboratory for Brain-inspired Intelligence Technology and Application (NEL-BITA), University of Science and Technology of China, Hefei 230027, China (e-mail:jfzhuang@mail.ustc.edu.cn; zlwang@ustc.edu.cn; wbkup@mail.ustc.edu.cn).}
	\thanks{Digital Object Identifier: 10.1109/TCSVT.2020.3037234}}

%
%

\markboth{IEEE Transactions on Circuits and Systems for Video Technology}%
{Jiafan \MakeLowercase{\textit{et al.}}: IEEE Transactions on Circuits and Systems for Video Technology}
%

\IEEEpubid{
	\begin{minipage}{\textwidth}
		\centering
		Copyright~\copyright~2020 IEEE. Personal use of this material is permitted. However, permission to use this \\ material for any other purposes must be obtained from the IEEE by sending an email to pubs-permissions@ieee.org.
	\end{minipage}
}



\newcommand{\etal}{\textit{et al}.}
\newcommand{\ie}{\textit{i}.\textit{e}.}
\newcommand{\eg}{\textit{e}.\textit{g}.}
\newcommand{\vs}{vs. }

\maketitle

\begin{abstract}
	Video semantic segmentation aims to generate an accurate semantic map for each frame in a video. For such a task, conducting per-frame image segmentation is generally unacceptable in practice due to high computation cost. To address this issue, many works perform the flow-based feature propagation to reuse the features of previous frames, which essentially exploits the content continuity of consecutive frames. 
	However, the estimated optical flow would inevitably suffer inaccuracy and then make the propagated features distorted. In this paper, we propose a distortion-aware feature correction method with the goal of improving video segmentation performance at a low price. Our core idea is to correct the features on distorted regions using the current frame while reserving the propagated features for other regions. In this way, a lightweight network is enough for achieving promising segmentation results. In particular, we propose to predict the distorted regions by utilizing the consistency of distortion patterns in images and features, such that the high-cost feature extraction from current frames can be avoided. We conduct extensive experiments on Cityscapes, CamVid, and UAVid, and the results show that our proposed method significantly outperforms previous methods and achieves the state-of-the-art performance on both segmentation accuracy and speed. Code and pretrained models are available at https://github.com/jfzhuang/DAVSS.
\end{abstract}

\begin{IEEEkeywords}
	Video semantic segmentation, feature propagation, distortion prediction, feature correction.
\end{IEEEkeywords}

%
\IEEEpeerreviewmaketitle

\section{Introduction}
\label{sec:introduction}
\IEEEPARstart{S}{emantic} segmentation is to assign each pixel in the scene a semantic class, which is currently an active research topic in computer vision. In recent years, image semantic segmentation has achieved unprecedented accuracy benefited from the great progress of deep convolutional neural networks (DCNN)~\cite{long2015fully} and various datasets (\textit{e.g.}, Cityscapes~\cite{Cordts_2016_CVPR}, CamVid~\cite{brostow2009semantic}, and UAVid~\cite{lyu2020uavid}).
However, many real-world applications have strong demands for fast and accurate video semantic segmentation, \textit{e.g.}, robotics~\cite{kostavelis2015semantic}, autonomous driving~\cite{teichmann2018multinet}, and video surveillance~\cite{Liu_2017_CVPR}.
Compared to images, videos involve much larger volume of data, and thus always require more efficient segmentation algorithms.

\begin{figure}[t]
	\begin{center}
		\includegraphics[width=1.0\linewidth]{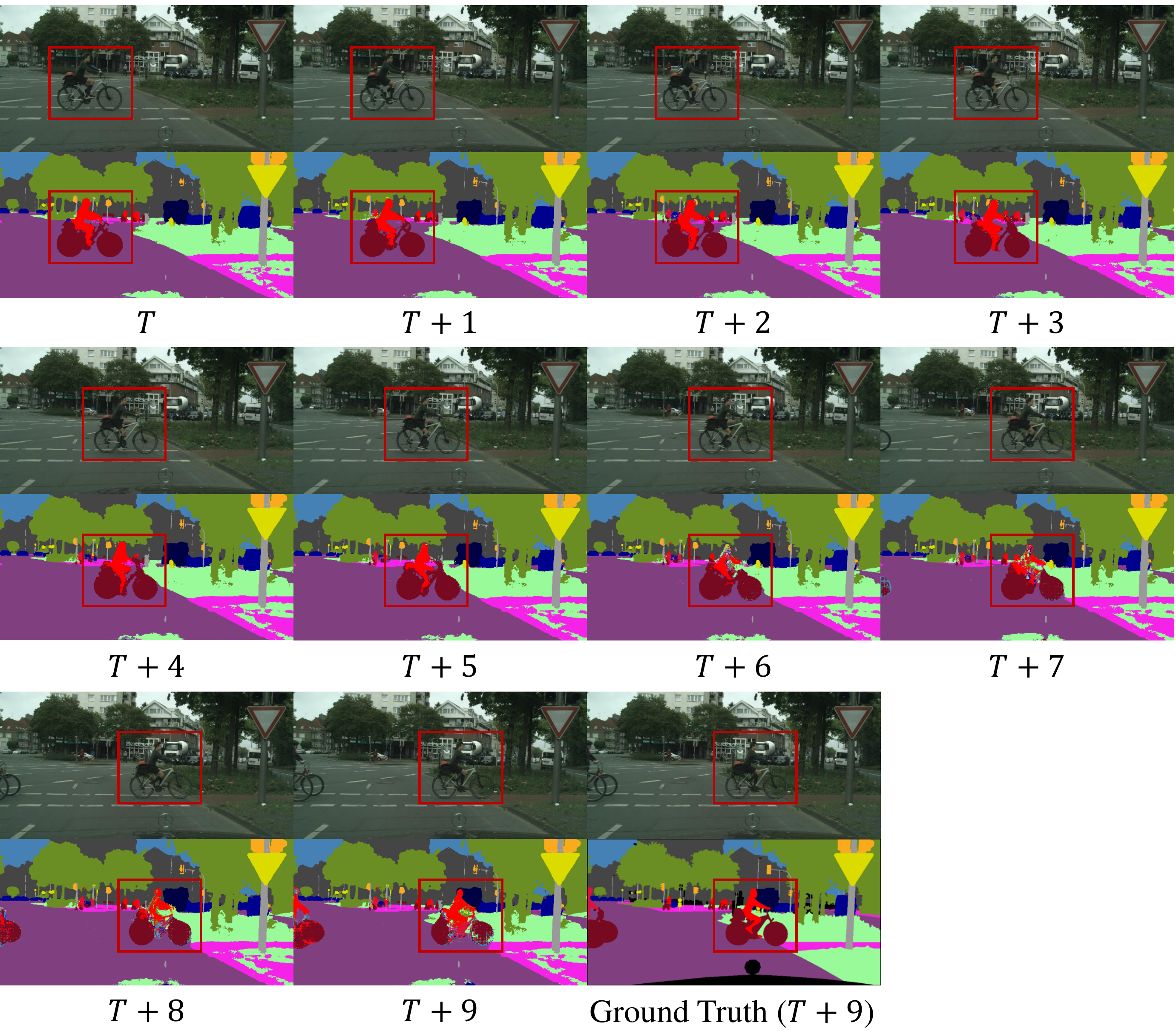}
	\end{center}
	\caption{\textbf{Illustration of distortion phenomenon in feature propagation.} The segmentation results of an example video produced by DFF~\cite{zhu2017deep} are demonstrated, where $T+i$ denotes the $i$-th frame from the key frame $T$. In particular, we also give the ground-truth segmentation of the frame $T+9$ for comparison. Red rectangles highlight the distorted regions caused by inaccurate optic flow estimation. Best viewed in color and zoom in.}
	\label{dff_distortion}
\end{figure}

A naive approach for video segmentation is to directly apply the image segmentation model in a per-frame way. But such deployment is generally unacceptable in practice due to too heavy computational burden. Actually, the consecutive frames of a video are often similar in a large portion of content, and it is unnecessary to reprocess every pixel of the video frame using an image segmentation model~\cite{xu2018dynamic}. Then an intuitive idea for video semantic segmentation is to reuse the features extracted from the previous frames when segmenting the current frame~\cite{zhu2017deep}. Naturally, the feature propagation is proposed to reduce the overall computational complexity.

\IEEEpubidadjcol

In recent years, some feature propagation based methods have been proposed for video semantic segmentation, \textit{e.g.}, DFF~\cite{zhu2017deep}, NetWarp~\cite{gadde2017semantic}, DVSNet~\cite{xu2018dynamic}, and Accel~\cite{jain2019accel}. These methods first compute the optical flow between the key frame and the current frame, and then produce the features of the current frame by propagating the features of the key frame under guidance of optical flow. Here the bilinear interpolation is usually used as the feature warping operator. The CNN-based optical flow estimation methods (\eg, FlowNet~\cite{dosovitskiy2015flownet,ilg2017flownet}, FlowNet2.0~\cite{ilg2017flownet}) are preferred since they are easy to be embedded into the video segmentation framework for end-to-end training. Evidently, the accuracy of optical flow estimation would determine the quality of propagated features and performance of semantic segmentation. 

Despite the great progress in the past decades, accurate optical flow estimation remains a challenging problem~\cite{liu2019selflow}. In particular, the occlusion caused by scene motion makes the optical flow estimation ill-posed since no visual correspondence exists for the occluded pixels~\cite{neoral2018continual}. When the inaccurate optical flow is used in feature propagation, the produced features would get distorted and incorrect segmentation results may be further generated. In addition, for small or slender areas of a single class (\eg, pedestrian, pole), a slight offset of predicted optical flow would cause sensible distortion, which is especially serious for long-distance propagation. We show the typical distortion phenomenon in Fig.~\ref{dff_distortion}. The distortion of feature propagation needs to be carefully tackled in video semantic segmentation.

\begin{figure}[t]
	\begin{center}
		\includegraphics[width=1.0\linewidth]{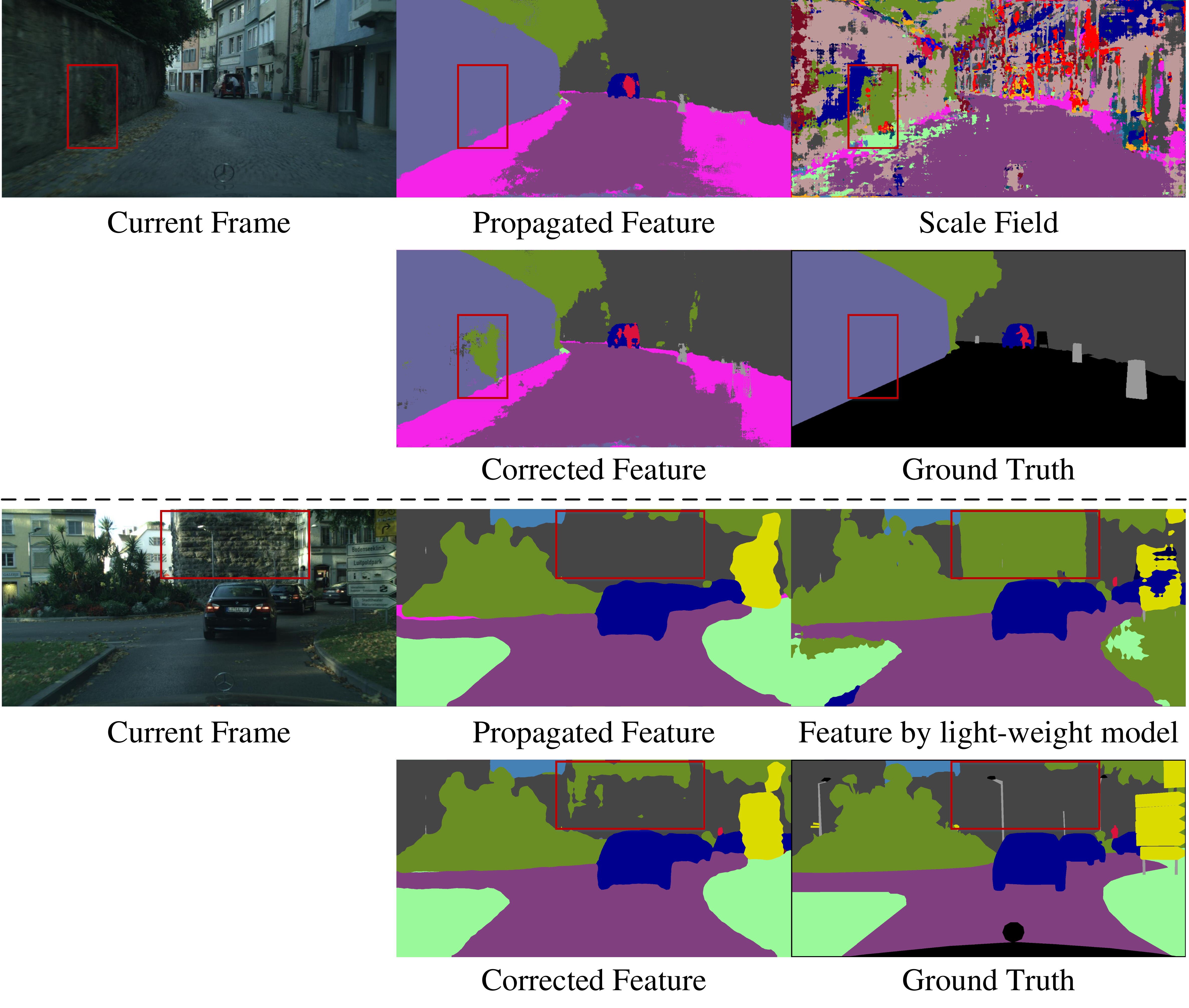}
	\end{center}
	\caption{\textbf{Visualization of false correction.} The propagated and corrected features are visualized and represented by their segmentation results. The upper case is from DFF and the blow one is from Accel50. Red rectangles highlight the areas corrected wrongly. Best viewed in color.}
	\label{overly_correction}
\end{figure}

\begin{figure}[t]
	\begin{center}
		\includegraphics[width=0.7\linewidth]{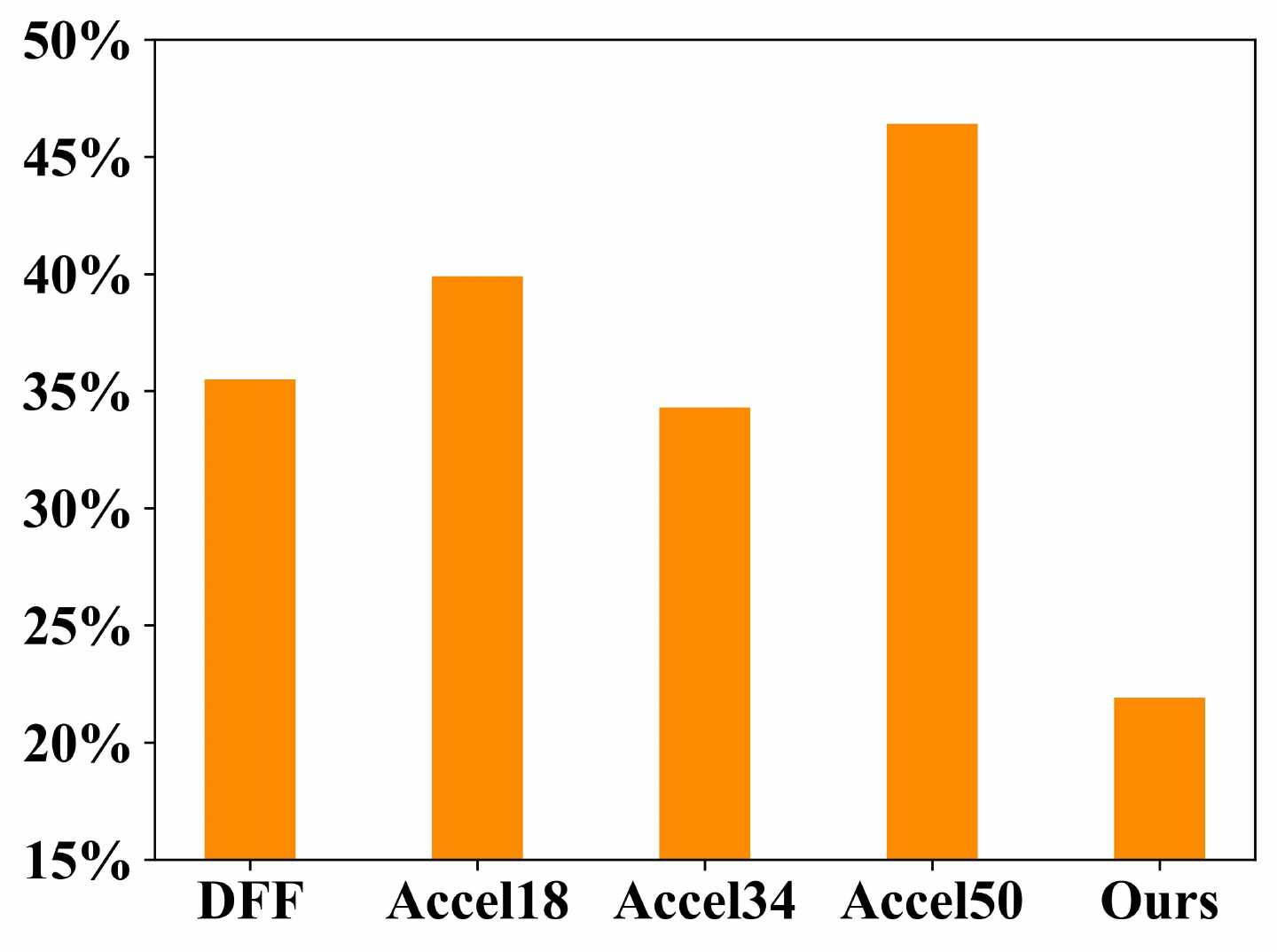}
	\end{center}
	\caption{\textbf{Statistics of false correction on Cityscapes val subset.} Here the ratio of the number of pixels wrongly and rightly rectified is particularly calculated for different methods. It can be seen that some correct propagated features would be wrongly rectified, and our proposed method can achieve the best result.  }
	\label{exp_correct_error}
\end{figure}

Some existing methods can alleviate feature distortion by modulating the propagated features. For example, DFF~\cite{zhu2017deep} attaches a scale field to optical flow estimation and adjusts the propagated features via element-wise multiplication. Accel~\cite{jain2019accel} proposes to extract features from the current frame with a lightweight model and then fuse the extracted and propagated features to perform semantic segmentation. However, these works equally treat every pixel of the current frame without distinguishing the quality of propagated features among different pixels. Consequently, the regions where the features are correctly propagated may be modulated to be wrong, namely, false correction may occur. We show typical cases of such a phenomenon in Fig.~\ref{overly_correction}. Besides, we check the ratio of the number of pixels wrongly and rightly rectified, and the statistics of different methods on \textit{Cityscapes val subset} are shown in Fig.~\ref{exp_correct_error}. Evidently, many correct features are wrongly rectified, even for models equipped with a heavy network (\eg, Accel50). 

\begin{figure}[t]
	\begin{center}
		\includegraphics[width=1.0\linewidth]{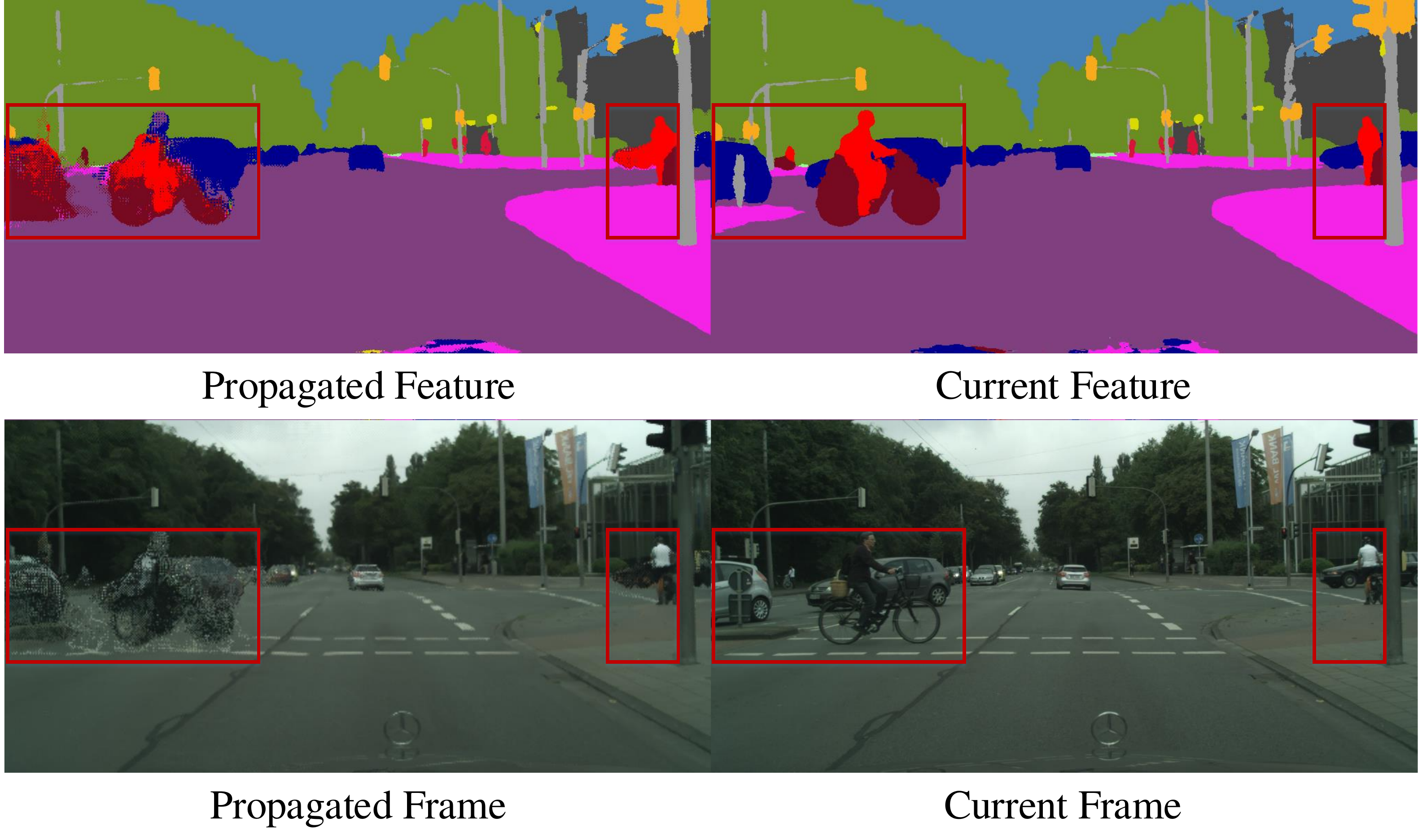}
	\end{center}
	\caption{\textbf{Illustration of distortion consistency for images and features.} We provide the segmentation results of the propagated and extracted features in the first row, and the propagated image from previous frame and current frame in the second row. Red rectangles highlight the main distorted regions. It can be seen that similar distortion patterns present for images and features. Best viewed in color and zoom in.}
	\label{distortion_pattern}
\end{figure}

In this work, we propose a novel distortion-aware feature correction method to rectify the propagated features, aiming at improving the accuracy of video semantic segmentation at a low price. Our key idea is to correct the features on the distorted regions while reserving the propagated features for other regions. With such a design, a lightweight network can be enough to perform the rectification. To this end, we need first to identify the distorted regions. An intuitive approach is to extract features from the current frame using an image segmentation model, and then get the misalignment regions by comparing the extracted and propagated features. However, extracting the features would involve too high computation cost. To tackle the issue, we propose to get the distorted regions through image comparison of video frames. Actually, the distortion is mainly caused by inaccurate optical flow, \textit{i.e.}, the distorted regions are essentially the regions where the optical flow is miscalculated. So we propose to concurrently propagate video frames with the same optical flow as in feature propagation, and then compare the propagated frame and current frame in the image space to get the distorted regions. Our proposed method essentially utilizes the consistency of the distortion patterns for images and features, as shown in Fig.~\ref{distortion_pattern}. Following this idea, we propose a very lightweight network to predict the distortion maps.

Then we propose a feature correction module (FCM) to perform distortion correction on the propagated features. Here the predicted distortion maps are utilized in two folds. First, we propose a CFNet to extract the correction cues from the current frame, and enforce it to focus on the distorted regions by applying the distortion map to the calculation of training loss. CFNet can be designed very lightweight since only the capacity to process the distorted regions is required. Second, FCM uses the distortion maps to identify the important regions on which the propagated features need to be rectified greatly by correction cues. Consequently, the correction cues from the current frame dominate the distorted regions while the propagated features dominate other regions. Finally, we conduct semantic segmentation on the corrected features.

The contributions of this work are summarized as 
\begin{itemize}
	\item We propose a novel distortion-aware feature correction method for video semantic segmentation, which can effectively boost the segmentation performance at a low price by focusing on the distorted regions. 
	\item We propose a lightweight network to predict the distorted regions of propagated features, which works in the image space and can effectively guide feature correction. 
	\item We experimentally evaluate the effectiveness of our proposed method, and the results on Cityscapes, CamVid, and UAVid demonstrate the superiority of our method to previous state-of-the-art methods, especially for long-distance feature propagation.
\end{itemize}

The rest of this paper is organized as follows. We review the related works on image and video semantic segmentation, optical flow estimation, and attention mechanism in Section~\ref{sec:related}. Section~\ref{sec:approach} provides the details of our approach, and Section~\ref{sec:exp} experimentally evaluates the proposed method. Finally, we conclude the work in Section~\ref{sec:conclusion}.

\section{Related Work} \label{sec:related}

\subsection{Image Semantic Segmentation}
Benefited from the rapid development of DCNN~\cite{iandola2016squeezenet, simonyan2014very, he2016deep, szegedy2015going, huang2017densely}, more and more semantic segmentation networks spring up. Specifically, the fully convolutional network (FCN)~\cite{long2015fully} firstly uses the convolutional layers to replace fully-connected layers, and better performance is achieved. Inspired by FCN, many extensions~\cite{zhao2017pyramid, wu2019wider, lin2017refinenet} have been proposed, which together advance image semantic segmentation. The dilated layers~\cite{chen2018deeplab, yu2015multi} are also introduced to replace the pooling layers, which can better balance the computational cost and size of receptive fields. In addition, \cite{chen2018deeplab, chen2014semantic, zheng2015conditional} propose to use the conditional random field (CRF) to refine the results of image segmentation. Recently, spatial pyramid pooling~\cite{he2015spatial} and atrous spatial pyramid pooling (ASPP)~\cite{chen2017rethinking, chen2018deeplab} are respectively used in PSPNet~\cite{zhao2017pyramid} and DeepLab~\cite{chen2018deeplab} to capture multi-scale contextual information. MPF~\cite{wang2018detecting} uses a new structural context descriptor and a self-weighted multiview clustering method for robust group detection. Priori s-CNNs~\cite{wang2017joint} learns priori location information at superpixel level and adopts a soft restricted MRF energy function to reduce over smoothness. CCNet~\cite{Huang_2019_ICCV} contains a criss-cross attention module to harvest the contextual information. HRNet~\cite{Sun_2019_CVPR} maintains the high resolution feature in the whole process and fuses multi-resolution features repeatedly for reliable and discriminative representations. SANet~\cite{Zhong_2020_CVPR} applies the pixel-group attention to capture spatial-channel inter-dependencies. Li \etal~\cite{Li_2020_CVPR} propose to generate data-dependent routes for adapting to the scale distribution of each image. Lin \etal~\cite{lin2020cross} propose to use skeleton representation to effectively bridge the synthesis and real domains and achieve comparable performance on multi-person part segmentation without any human-annotated labels. Ca-crfs Net~\cite{ji2020encoder} introduces cascaded CRFs into the decoder to learn boundary information and enhance the ability of object boundary location. The great progress of image semantic segmentation offers foundation for video semantic segmentation.

\begin{figure*}[t]
	\begin{center}
		\includegraphics[width=0.8\linewidth]{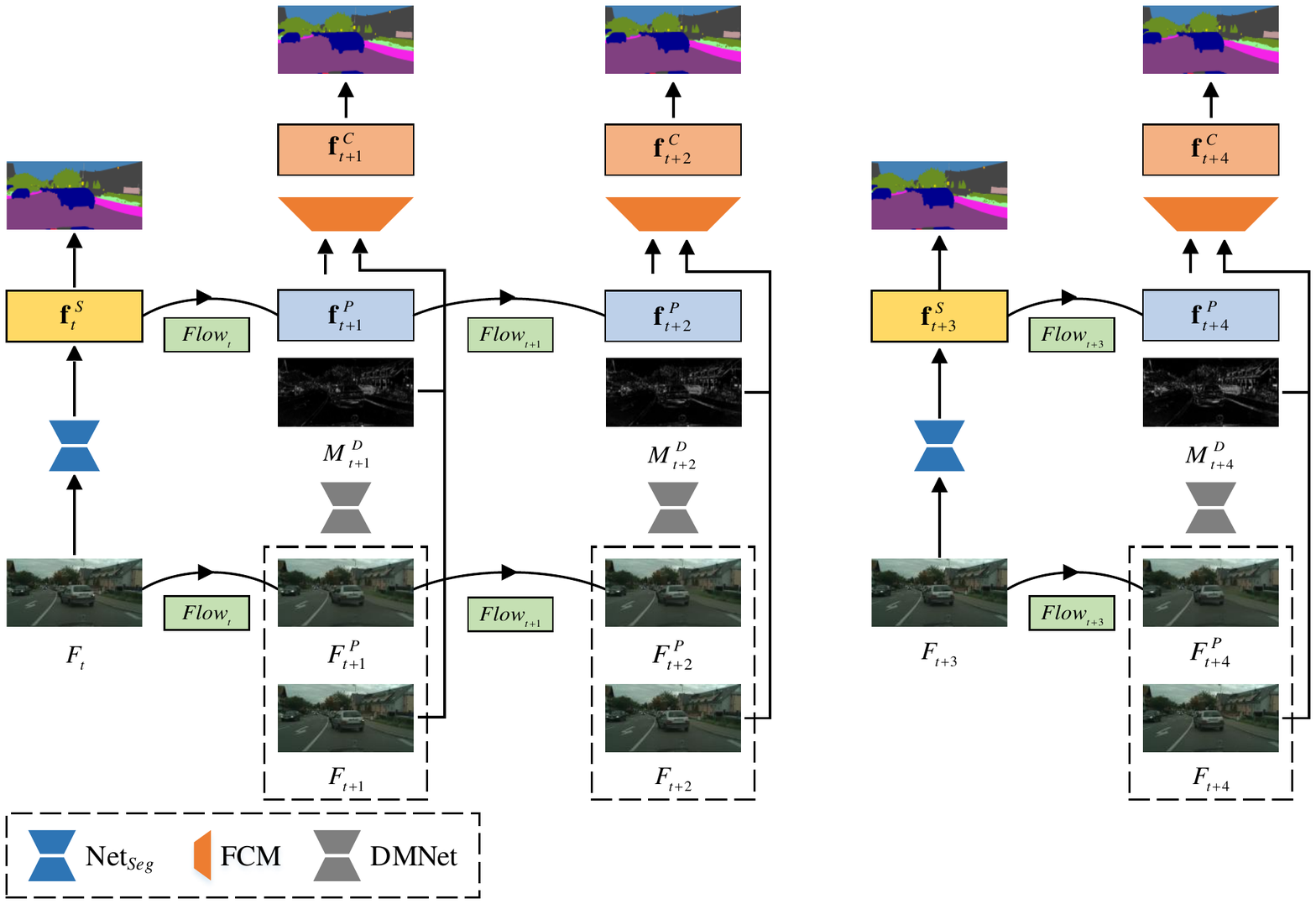}
	\end{center}
	\caption{\textbf{Framework of our proposed approach}. $F$ and $F^{P}$ represent the original video frame and propagated one from previous frame, respectively. Particularly, the frames $F_{t}$ and $F_{t+3}$ are selected as the key frames for illustration. In real deployment, the key frames can be selected by a fixed-interval schedule like in~\cite{zhu2017deep} or an adaptive schedule like in~\cite{xu2018dynamic} and~\cite{li2018low}. For the key frames, the feature $\mathbf{f}^{S}$ is extracted via an image segmentation network Net$_{seg}$. For the non-key frames, the propagated feature $\mathbf{f}^{P}$ is first produced through frame-by-frame propagation, and then is rectified into $\mathbf{f}^{C}$ by feature correction module (FCM) that combines the correction cues extracted from the current frame under the guidance of the distortion map $M^{D}$. Here $M^{D}$ is predicted by a lightweight network DMNet taking as input the propagated and current frames. Best viewed in color.}
	\label{framework}
\end{figure*}

\subsection{Optical Flow Estimation}
Optical flow is a representative pattern describing the apparent motion of objects in the video. Optical flow estimation is a fundamental task in the video analysis domain. Classical variational approaches formulate optical flow estimation as an energy minimization problem~\cite{horn1981determining,anguita2009optimization}. Such methods are effective for small motion, but tend to fail when displacements are relatively larger. Recent works use convolutional neural networks (CNNs) to improve sparse matching by learning an effective feature embedding~\cite{dosovitskiy2015flownet,ilg2017flownet,sun2018pwc,zhai2019optical}. 

Although current methods can generate satisfactory optical flow in most common cases, it is still a challenging problem to calculate accurate optical flow for occlusion areas. Most methods detect occlusion by consistency check on the estimated forward and backward optical flow~\cite{chen2016full,sundaram2010dense}, and then extrapolating the occluded areas. But the used optical flow would be adversely affected by the occlusion. Evidently, the propagated features under the guidance of inaccurate optical flow would be severely distorted, especially for occlusion areas. 

Actually,  most video semantic segmentation methods prefer the current state-of-the-art CNN networks for predicting the optical flow~\cite{dosovitskiy2015flownet,ilg2017flownet,sun2018pwc,ranjan2017optical} because they are easily embedded for end-to-end training. However, these methods do not explicitly deal with occlusion, and consequently video segmentation would suffer from severe feature distortion. Thus how to deal with the feature distortion  efficiently and effectively is crucial for the optical flow based video segmentation methods.

\subsection{Attention Mechanism}
Attention mechanism has been widely used in computer vision. It can effectively increases the representational power of neural networks by selectively weighting the feature maps. For example, ~\cite{hu2018squeeze} proposes a squeeze-and-excitation (SE) module to adaptively recalibrate channel-wise feature responses by explicitly modeling interdependencies between channels. OCNet~\cite{yuan2018ocnet} and DANet~\cite{fu2019dual} utilize self-attention mechanism to harvest the contextual information. Chen~\etal~\cite{chen2018reverse} propose the reverse attention to guide side-output residual learning in a top-down manner. Chen~\etal~\cite{chen2018embedding} propose a visual attention mechanism which can bridge high-level semantic information to help the shallow layers locate salient objects and filter out noisy response in the background region. BiANet~\cite{zhang2020bilateral} introduces a bilateral attention module to focus on the foreground region with a gradual refinement style and recover potentially useful salient information in the background region. Fan~\etal~\cite{fan2020pranet} proposes a parallel reverse attention network to aggregate the features in high-level layers and mine the boundary cues using the reverse attention module. In this paper, we propose to only focus on the distorted regions of propagated features under the guidance of predicted distortion maps. 

\subsection{Video Semantic Segmentation}
Different from static images, videos embody useful temporal information that can be exploited.
So many previous works focus on modeling cross-frame relations to integrate the information from different frames to boost the semantic segmentation accuracy. STFCN~\cite{fayyaz2016stfcn} utilizes a spatial-temporal LSTM over per-frame CNN features. Nilsson and Sminchisescu~\cite{nilsson2018semantic} propose to use the gated recurrent units to propagate semantic labels. Gadde~\etal~\cite{gadde2017semantic} propose to fuse the features warped from the key frame and those from the current frame. V2V~\cite{tran2016deep} utilizes a 3D CNN to perform a voxel-level prediction. Wang~\etal~\cite{wang2019superpixel} propose a metadata-based global projection model with the coordinate transformation to estimate motion information between frames.

On the other hand, many works reduce the overall computation cost of video semantic segmentation by utilizing the content continuity of consecutive frames. Clockwork Net~\cite{shelhamer2016clockwork} updates different levels of feature maps with different frequencies. DFF~\cite{zhu2017deep} estimates the optical flow fields from the key frame to other frames and then propagates the high-level features using the predicted optic flows. DVSNet~\cite{xu2018dynamic} builds a decision model to dynamically choose the key frames, which can achieve better balance between quality and efficiency. Li~\etal~\cite{li2018low} propose spatially variant convolution to adaptively fuse the features over time. Accel~\cite{jain2019accel} proposes a reference branch to extract high-quality segmentation from the key frames and an update branch to efficiently extract low-quality segmentation from the current frames, and then fuses them to improve the segmentation accuracy. TDNet~\cite{hu2020temporally} distributes several sub-networks over sequential frames and then recomposes the extracted features for segmentation via an attention propagation module.

A related task to video semantic segmentation is video object segmentation. Several works~\cite{wang2005interactive,li2005video,price2009livecut} reduce the structural complexity of the graphical model with spatio-temporal superpixels. Chen~\etal~\cite{chen2019motion} propose a two-stage framework of integrating motion and appearance cues for foreground object segmentation in unconstrained videos. Liu~\etal~\cite{liu2020guided} propose a guided co-segmentation network to simultaneously incorporate the short-term, middle-term, and long-term temporal inter-frame relationships.

In this work, we follow the route of feature propagation for video semantic segmentation. Different from previous works that equally treat every pixel of a video frame, we propose to focus on the distorted regions when rectifying the propagated features. In this way, the semantic segmentation results can be more effectively enhanced, and the used network can be more lightweight for high efficiency.

\section{Our Approach} \label{sec:approach}

In this work, we try to boost the accuracy of video semantic segmentation on the non-key frames effectively and efficiently under the framework of optical flow based feature propagation. To this end, we propose a distortion-aware feature correction method, and the core idea is to correct the features on the distorted regions while reserving the propagated features for other regions.  For such an idea, we need to design an elegant solution to address the following issues, namely, 1) how to identify the distorted regions, 2) how to extract the correction cues from the current frame, and 3) how to effectively perform feature correction. 
In the following, we first introduce the framework of our proposed method. Then we elaborate on two main components of the proposed method: distortion map prediction and feature correction. Finally, we provide training details of our proposed network. 

\subsection{Framework}

The framework of our proposed approach is illustrated in Fig.~\ref{framework}, where semantic segmentation is performed on the feature of each frame individually.
To be specific, each of the video frames is treated as the key or non-key frame. For the key frames, we directly conduct image semantic segmentation to get the results using an off-the-shelf network, and the intermediate features are propagated to subsequent non-key frames. In particular, we propagate the features in a frame-by-frame way during inference. That is, the feature of the current frame is first produced by propagating that of the previous frame, in which the predicted optical flow is used as the guidance and the bilinear interpolation is usually adopted as the warping operator. Along with feature propagation, we also propagate the video frame with the same optical flow, resulting in the propagated frame. For the non-key frames, we first feed the propagated frame and current frame into our proposed distortion map network (DMNet) to predict a distortion map, which actually represents the distortion pattern of the propagated feature. Then we use the current frame to rectify the propagated feature under the guidance of the predicted distortion map. We complete such feature rectification in the proposed feature correction module (FCM). Finally, we conduct semantic segmentation on the corrected feature to get the segmentation result of the current frame.

The key components of our proposed framework are the distortion map prediction and feature correction. In our implementation, we particularly adopt DeepLabv3+~\cite{chen2018encoder} as the image semantic segmentation model due to its great performance in both accuracy and efficiency. The modified FlowNet2-S~\cite{ilg2017flownet} is used for optical flow estimation. 

\begin{figure}[t]
	\begin{center}
		\includegraphics[width=1.0\linewidth]{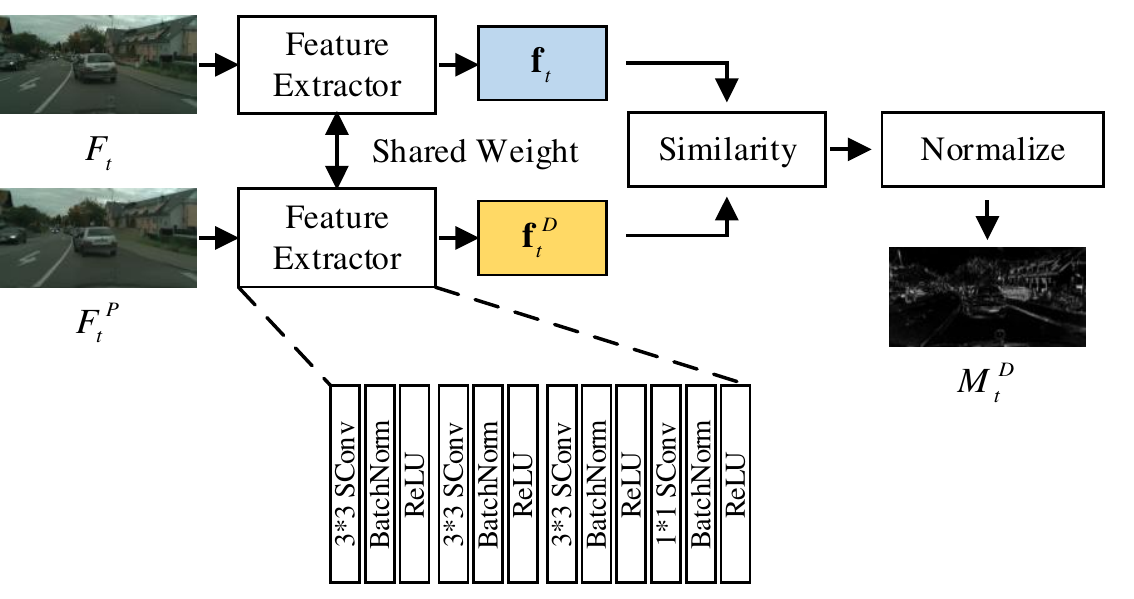}
	\end{center}
	\caption{\textbf{Illustration of our proposed DMNet}. Following the design of siamese networks, DMNet takes in the propagated frame and current frame to perform feature extraction and similarity computation.}
	\label{DMNet}
\end{figure}

\begin{figure}[t]
	\begin{center}
		\includegraphics[width=0.9\linewidth]{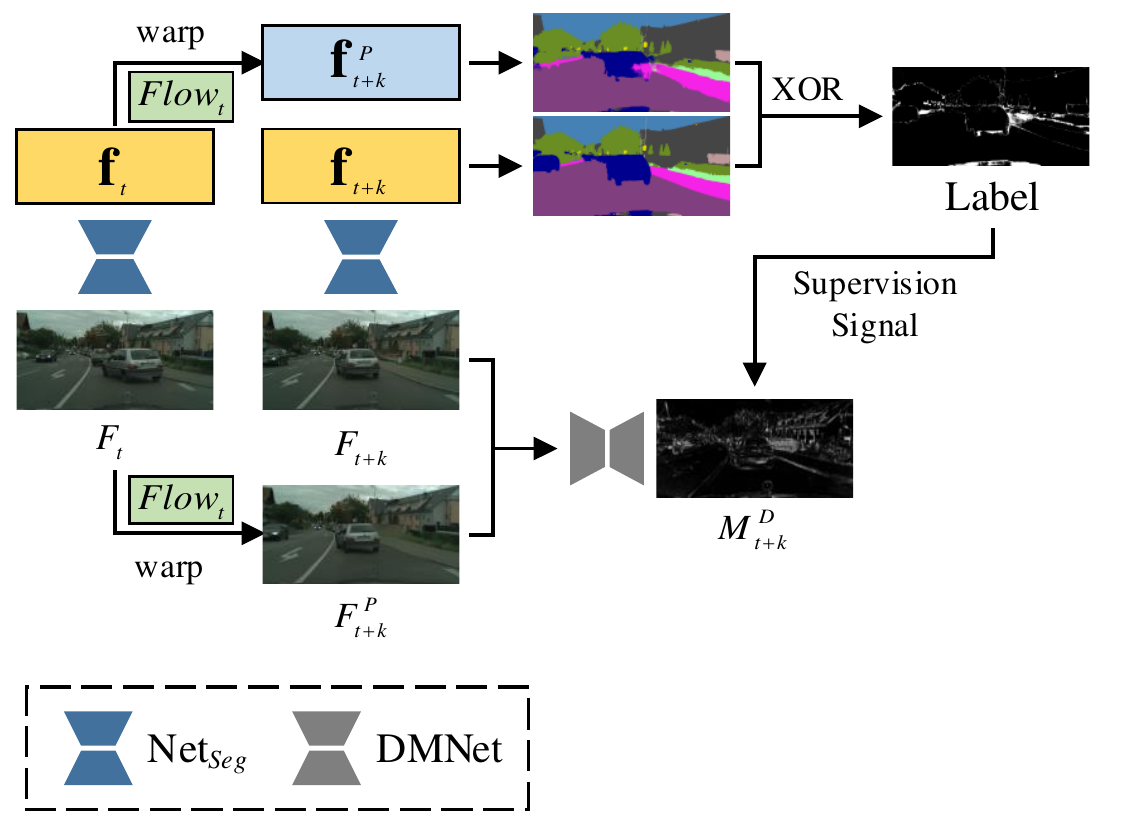}
	\end{center}
	\caption{\textbf{Illustration of training DMNet}. $Flow_{t}$ represents the predicted optical flow from $F_{t}$ to $F_{t+k}$. The distortion map for training DMNet is obtained by calculating the difference between the segmentation results of the propagated features and current frame.}
	\label{DMNet_training}
\end{figure}

\subsection{Distortion Map Prediction}

In this work, we propose a distortion map network (DMNet) to predict the distorted area of propagated features. Here we do not extract any high-level feature from the current frame since it involves too high computation cost. Instead, we compare the propagated frame and the current frame to get the distorted regions, which actually exploits the consistency of distortion pattern for images and features. 
To be specific, we follow the design of siamese networks to build DMNet that calculates the difference between the propagated frame and the current frame, as shown in Fig.~\ref{DMNet}. To achieve high computational efficiency, the feature extractor is designed to only comprise four separable convolutional layers interlaced with BatchNorm and ReLU layers. Consequently, the involved computation cost is nearly negligible. Then we can calculate the cosine similarity of two features, resulting a similarity map ${S}$. Formally, let $\mathbf{f}_{t}$ and $\mathbf{f}^{D}_{t}$ denote the features from the current frame $F_{t}$ and propagated frame $F^{P}_{t}$, respectively. Then

\begin{equation}
S_{t}(p)=\langle\bar{\mathbf{f}}_{t}(p), \bar{\mathbf{f}}^{D}_{t}(p)\rangle=\bar{\mathbf{f}}^{D}_{t}(p) \bar{\mathbf{f}}^{T}_{t}(p),
\end{equation}
where $p$ denotes the spatial position, $\bar{\mathbf{f}}=\mathbf{f}/\|\mathbf{f}\|_{2}$ denotes the $\ell_{2}$-normalized feature, and $\bar{\mathbf{f}}^{T}$ is the transpose of $\bar{\mathbf{f}}$. Obviously, the distorted regions would have lower value on the similarity map. To obtain the distortion map $M^D_{t}$, we normalize the similarity map as 
\begin{equation}
M^D_{t}=(-S_{t}+1)/2.
\end{equation}

In our implementation, we use the supervised learning to train DMNet, as shown in Fig.~\ref{DMNet_training}. Here the ground truth of distortion maps is obtained by calculating the difference of segmentation results between the propagated feature and extracted feature from the current frame. To be specific, we propagate the feature of $F_{t}$ to $F_{t+k}$ to get the propagated feature $\mathbf{f}^{P}_{t+k}$, where $F_{t+k}$ denotes the $k$-th frame from the frame $F_{t}$. Then we get the segmentation result of $\mathbf{f}^{P}_{t+k}$ via an argmax operation. Meanwhile, we get the segmentation result of $F_{t+k}$ using the image segmentation model. 
Finally, we produce the distortion map  for DMNet by applying the XOR operator on two segmentation results.

\begin{figure}[t]
	\begin{center}
		\includegraphics[width=1.0\linewidth]{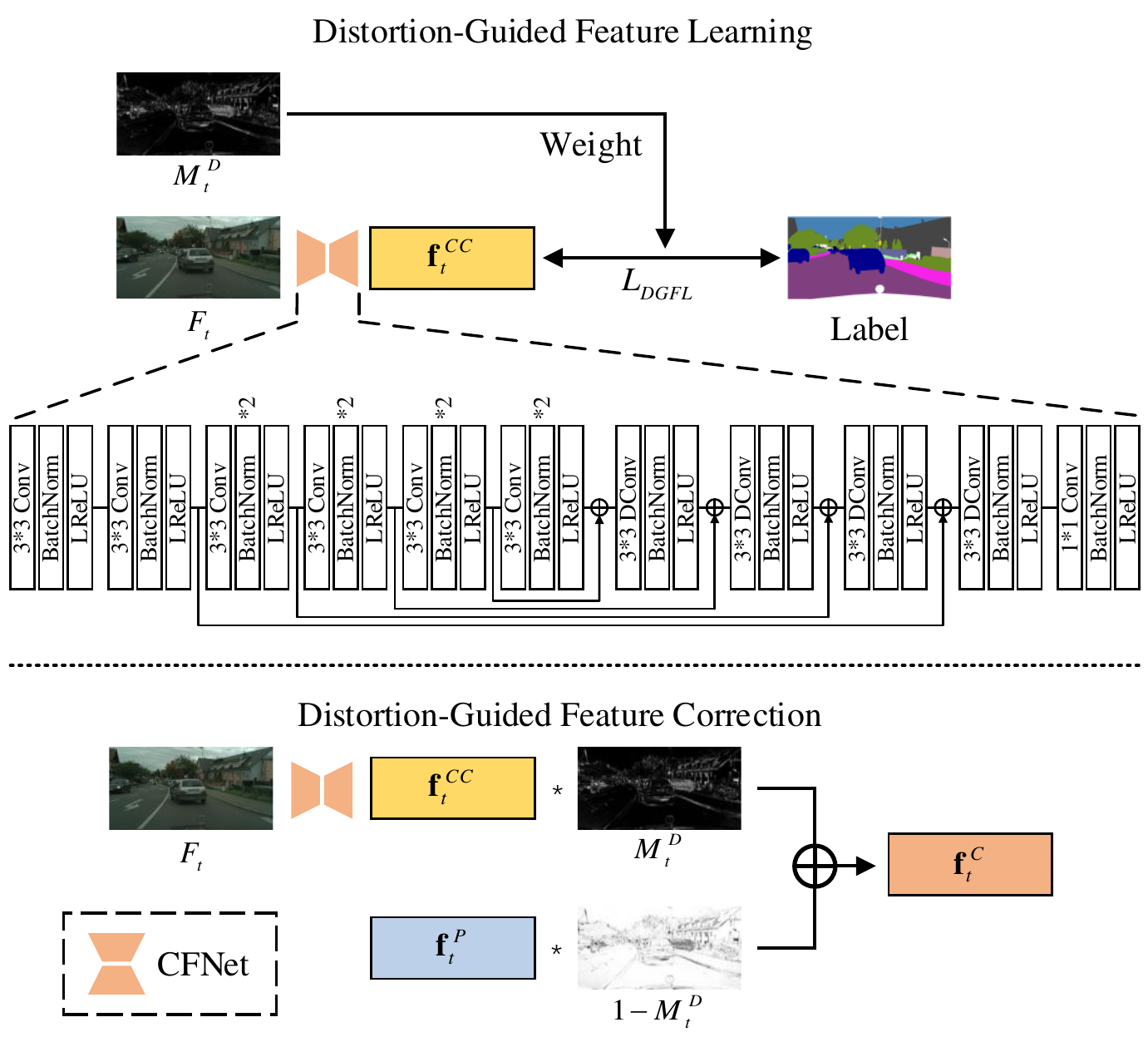}
	\end{center}
	\caption{\textbf{Illustration of feature correction module (FCM)}. Here the predicted distortion map $M^{D}_{t}$ is used in two folds. First, it guides the training of the network to extract correction cues from the current frame by weighting the loss of different regions. Second, it determines how to fuse the propagated feature $\mathbf{f}^{P}_{t}$  and extracted correction cue $\mathbf{f}^{CC}_{t}$ to get the corrected feature $\mathbf{f}^{C}_{t}$. Best viewed in color.}
	\label{FCM}
\end{figure}

\begin{figure}[t]
	\begin{center}
		\includegraphics[width=1.0\linewidth]{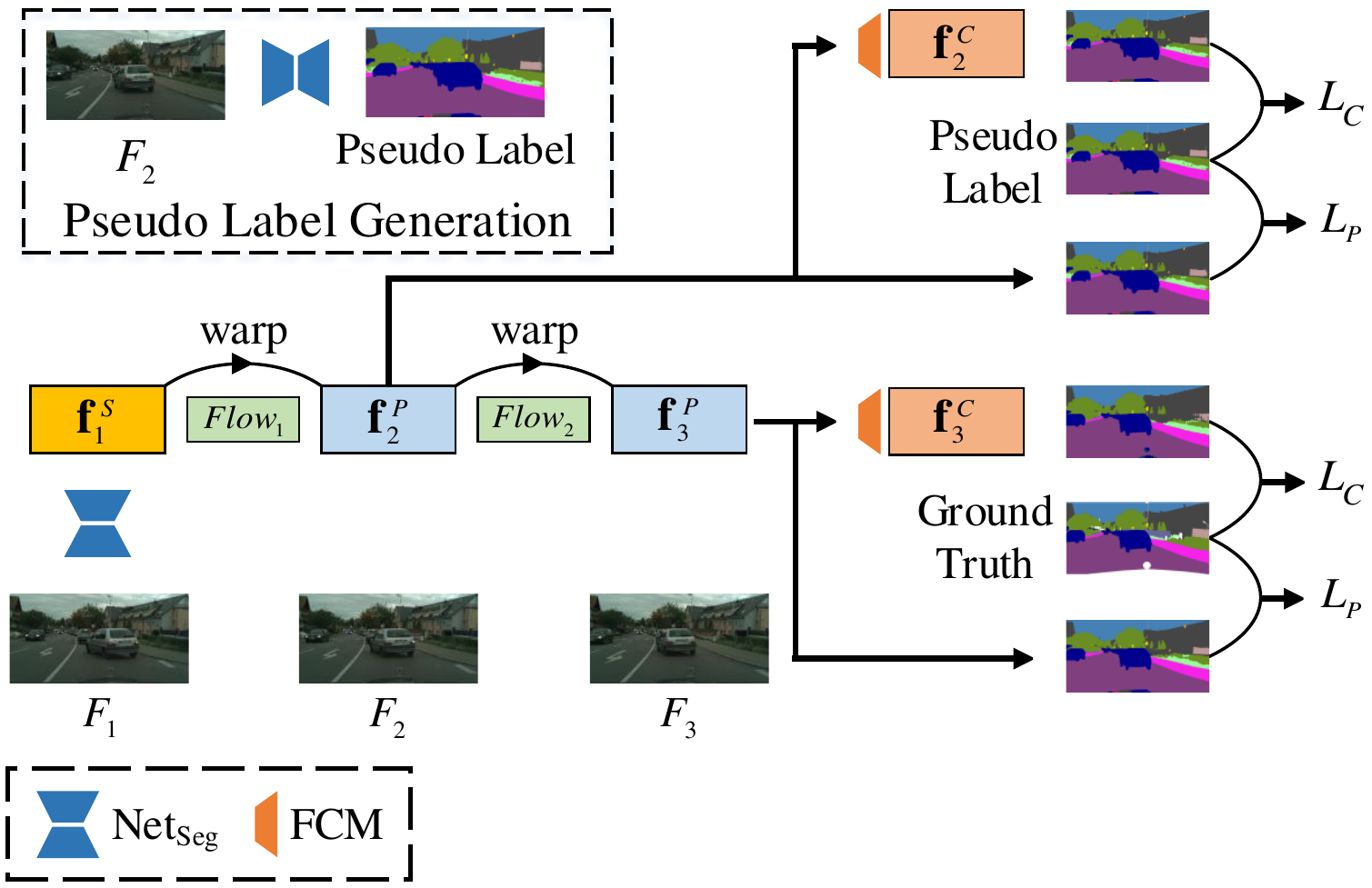}
	\end{center}
	\caption{\textbf{Illustration of training strategy}. We propose dual deep supervision (DDS) to improve the training of the network. Here $L_{P}$ and $L_{C}$ denote the loss calculated for feature propagation and correction, respectively. Best viewed in color.}
	\label{framework_training}
\end{figure}

\subsection{Feature Correction Module}

In this work, we explicitly rectify the propagated feature using the information of the current frame.  And we propose a feature correction module (FCM) to complete such feature rectification. Specifically, we consider two key goals of video semantic segmentation: high segmentation accuracy and low computation cost. To compromise two goals, we particularly propose to utilize the predicted distortion map to guide the feature correction.
In FCM, we utilize the distortion map in two folds, as shown in Fig.~\ref{FCM}. 

First, it is used to guide the extraction of correction cues from the current frame. Here it is expected that the correction cue can produce accurate  segmentation results over distorted regions and meanwhile the used network is lightweight enough for efficient computation. To this end, we propose CFNet in this work that mainly consists of ten convolutional layers interlaced with batchnorm and LReLU layers for feature encoding and four deconvolutional layers interlaced with LReLU layer for feature decoding. Then we use the distortion map to weight the \textit{CrossEntropy} loss when training CFNet, namely, distortion-guide feature learning (DGFL) is constructed. The loss function is
\begin{equation}
L_{DGFL}=-\frac{1}{HW}\sum_{h \in H,w \in W}M^{D}_{t}(h,w)\log{p_{t}^{CC}(h,w)},
\end{equation}
where $M^{D}_{t}$ is the distortion map and $p_{t}^{CC}$ is the predicted probability towards the ground truth from $\mathbf{f}_{t}^{CC}$. Through such loss weighting, CFNet would pay more attention on the distorted regions than others, and a lightweight model is enough to effectively extract discriminative features.

Second, the distortion map is used to determine how to rectify the propagated feature. Let $\mathbf{f}^{P}_{t}$ denote the propagated feature from the key frame to the current frame $F_{t}$,  $\mathbf{f}^{CC}_{t}$  be the extracted correction cue, and $\mathbf{f}^{C}_{t}$ be the corrected feature. 
FCM adopts the weighted sum to perform the feature correction, in which the features on the distorted regions are mainly rectified,  namely, distortion-guided feature correction (DGFC) is constructed. Then we have
\begin{equation}
\mathbf{f}^{C}_{t}=\mathbf{f}^{P}_{t}\odot(1-M^{D}_{t})+\mathbf{f}^{CC}_{t}\odot M^{D}_{t},
\end{equation}
where $\odot$ represents the spatially element-wise multiplication.
It can be seen that the features on the distorted regions are dominated by the correction cue $\mathbf{f}^{CC}_{t}$ while the features on other regions are dominated by the propagated features $\mathbf{f}^{P}_{t}$.

\subsection{Training Strategy}

Here we explain the training strategy of our proposed method, which is illustrated in Fig.~\ref{framework_training}. Before elaborating on the details, we briefly introduce the training procedure~\cite{zhu2017deep} widely used in previous works. For video semantic segmentation, let ($F_{1}$, $F_{3}$, GT) denote one training sample, where $F_{1}$ and $F_{3}$ are the key frame and current frame respectively, and GT is the segmentation ground truth of $F_{3}$. During training, $F_{1}$ is fed into the image segmentation model to extract features, and meanwhile the optical flow between $F_{1}$ and $F_{3}$ is estimated with FlowNet. Then the extracted features are propagated to $F_{3}$, and the \textit{CrossEntropy} loss at $F_3$ is calculated to train networks. In practice, $F_{1}$ is randomly selected from a $10$ frames video clip and $F_{3}$ is always the last one with ground truth, which can enrich the diversity of training samples.

However, the above training procedure may be unstable due to inaccuracy of optical flow estimation, especially for long-distance propagation (\eg, larger than $5$ frames). In this work, we propose \emph{dual deep supervision} (DDS) to improve network training by providing more supervision. Specifically, we add an intermediate frame for each training sample, denoted by $F_{2}$, to reduce the propagation distance, and meanwhile impose the supervision signal on $F_{2}$. Note that our method propagates the features frame-by-frame in the inference phase, and thus two-warp operation in the training phase is more appropriate than the original one.

In our experiments,  $F_{2}$ is randomly selected to ensure the diversity of training samples. To be specific, we extract the features of $F_{1}$, and conduct feature propagation twice ($F_1$ $\rightarrow$ $F_2$ $\rightarrow$ $F_3$).  Then we produce the pseudo label of $F_2$ using the image segmentation model for more supervision. Actually, using the pseudo label has been a natural and popular way to improve the segmentation quality in domain adaptation~\cite{zou2018unsupervised} and semi-supervised learning~\cite{hungadversarial}. Finally, we use the generated pseudo label and ground truth to supervise both the feature propagation and correction procedures on $F_{2}$ and $F_{3}$, as shown in Fig.~\ref{framework_training}. In particular, the propagation loss $L_{P}$ works on the warped features $\mathbf{f}_{t}^{P}$ for improving the quality of optical flow, and the correction loss $L_{C}$ works on the rectified features $\mathbf{f}_{t}^{C}$ for enhancing the ability of feature correction. The two losses can be written as :
\begin{equation}
L_{P}=-\frac{1}{HW}\sum_{h \in H,w \in W}\log{p^{P}_{t}(h,w)},
\end{equation}
\begin{equation}
L_{C}=-\frac{1}{HW}\sum_{h \in H,w \in W}\log{p^{C}_{t}(h,w)},
\end{equation}
where $p^{P}_{t}$ and $p^{C}_{t}$ are the predicted probability towards the ground truth from $\mathbf{f}^{P}_{t}$ and $\mathbf{f}^{C}_{t}$, respectively.
Taking the loss of feature learning in FCM $L_{DGFL}$, our final loss for one single frame is
\begin{equation}
L=(L_{P}+L_{C}+L_{DGFL})/3.
\end{equation}

Our method consists of four main components, \ie, Net$_{seg}$, FlowNet, DMNet, and CFNet. Here we explain how they are trained. Net$_{seg}$ is pretrained on ImageNet and then finetuned on a particular segmentation dataset (\eg, Cityscapes, CamVid, and UAVid). DMNet is trained with the generated ground-truth distortion maps. Net$_{seg}$ and DMNet would keep fixed in the following training procedure. FlowNet is pretrained on the synthetic Flying Chairs dataset~\cite{dosovitskiy2015flownet} and then jointly trained with the randomly initialized CFNet by following the proposed DDS training strategy.  For each step of training, we adopt the Adam optimizer~\cite{kingma2014adam} with $\beta_{1}=0.9$ and $\beta_{2}=0.99$. The learning rate is set to $10^{-4}$ for the first 50 epochs and then fixed to $10^{-5}$ for the rest 50 epochs.

\begin{table}[t]
	\caption{Calculation of FLOPs for different operators. $H_{i}$, $W_{i}$, and $C_{i}$ are the height, width, and channel of the input feature map, and $H_{o}$, $W_{o}$, and $C_{o}$ correspond to the output feature map. $K_{h}$ and $K_{w}$ represent the size of convolutional kernel.}
	\begin{center}
		\renewcommand{\arraystretch}{1.3}
		\begin{tabular}{l|c}
			\hline
			Layer             		& FLOPs											\\
			\hline
			Convolution          	& $2H_{o}W_{o}(C_{i} K_{h}K_{w} + 1)C_{o}$  	\\
			Bilinear Upsampling 	& $11H_{o}W_{o}C_{o}$							\\
			Batch Normalization 	& $2H_{i}W_{i}C_{i}$							\\
			ReLU or LReLU 			& $H_{i}W_{i}C_{i}$								\\
			\hline
		\end{tabular}
	\end{center}
	\label{flops}
\end{table}

\section{Experiment} \label{sec:exp}
In this section, we experimentally evaluate our proposed method on three challenging datasets, namely, Cityscapes~\cite{cordts2016cityscapes}, CamVid~\cite{brostow2009semantic}, and UAVid~\cite{lyu2020uavid}, and compare it with some state-of-the-art methods. We conduct all of the experiments on the NVIDIA GTX 1080Ti GPUs. 

\subsection{Datasets}
\textbf{Cityscapes}~\cite{cordts2016cityscapes} is a popular dataset in semantic segmentation and autonomous driving domain. It focuses on semantic understanding of urban street scenes. The training and validation subsets contain $2,975$ and $500$ video clips, respectively, and each video clip contains $30$ frames. The $20${th} frame in each clip is annotated by pixel-level semantic labels with $19$ categories.

\textbf{CamVid}~\cite{brostow2009semantic} also focuses on the semantic understanding of urban street scenes, but it contains less data than Cityscapes. It only has $701$ color images with annotations of $11$ semantic classes. CamVid is divided into the trainval set with $468$ samples and test set with $233$ samples.  All samples are extracted from driving videos captured at daytime and dusk, and have pixel-level semantic annotations. Each CamVid video contains $3,600$ to $11,000$ frames at a resolution of $720\times960$.

\textbf{UAVid}~\cite{lyu2020uavid} is a high-resolution Unmanned Aerial Vehicle (UAV) semantic segmentation dataset, which brings new challenges, including large scale variation, moving object recognition and temporal consistency preservation. The training and validation subsets contain $20$ and $7$ video clips, respectively, and each video clip contains $900$ frames at a resolution of $2160\times3840$. Every $100$ frames in each clip are annotated by pixel-level semantic labels with $8$ categories.

\begin{figure}[t]
	\begin{center}
		\includegraphics[width=1.0\linewidth]{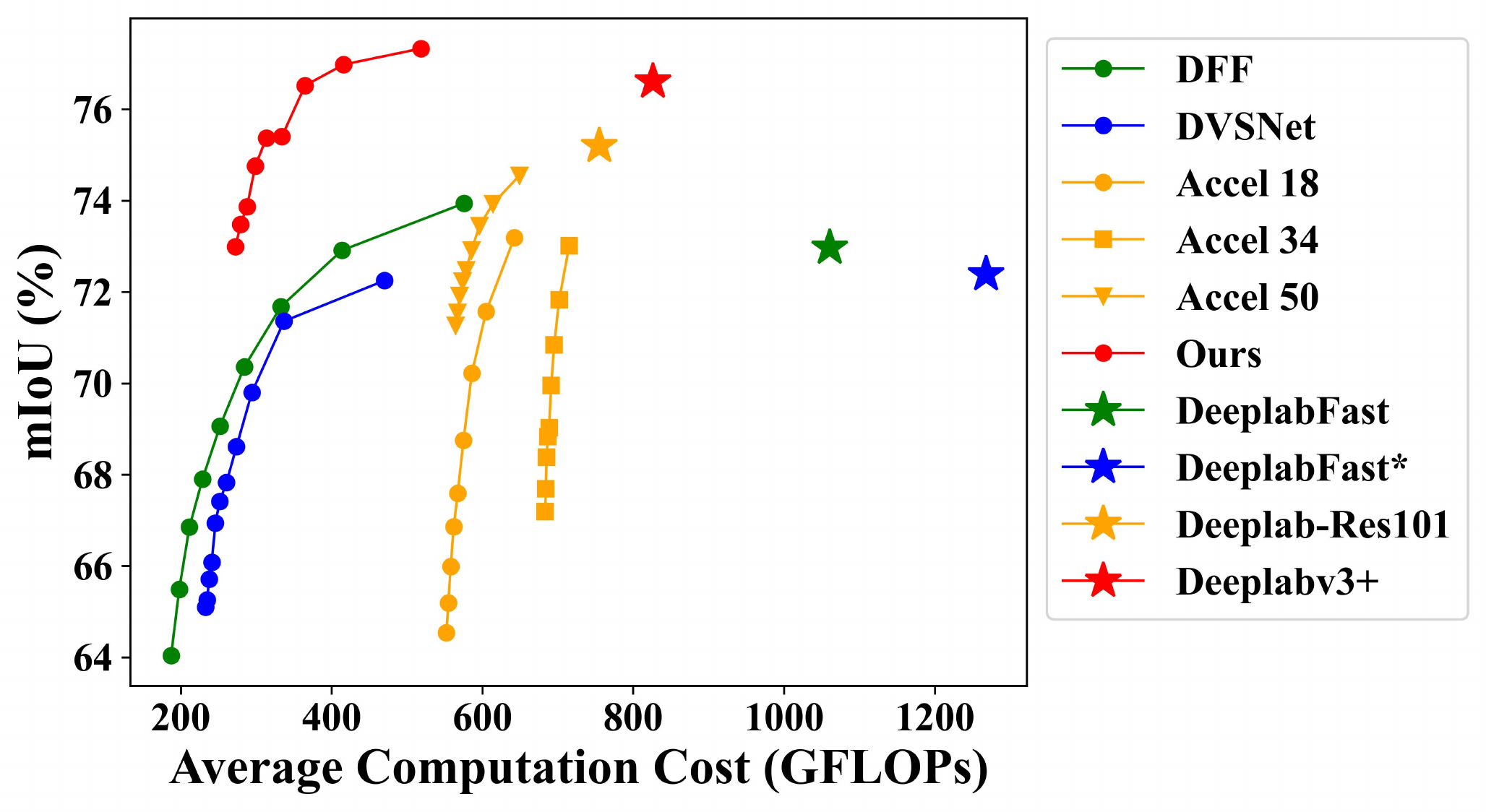}
	\end{center}
	\caption{\textbf{Performance comparison of different methods on Cityscapes val subset with CCA Curve.} Here $\bigstar$ denotes the results of per-frame image segmentation model.  In particular, ''DeeplabFast*'' represents the segmentation model used in DVSNet, which processes the regions of frame multiple times and thus has higher computation cost. Best viewed in color.}
	\label{exp_cityscapes}
\end{figure}

\begin{figure}[t]
	\begin{center}
		\includegraphics[width=1.0\linewidth]{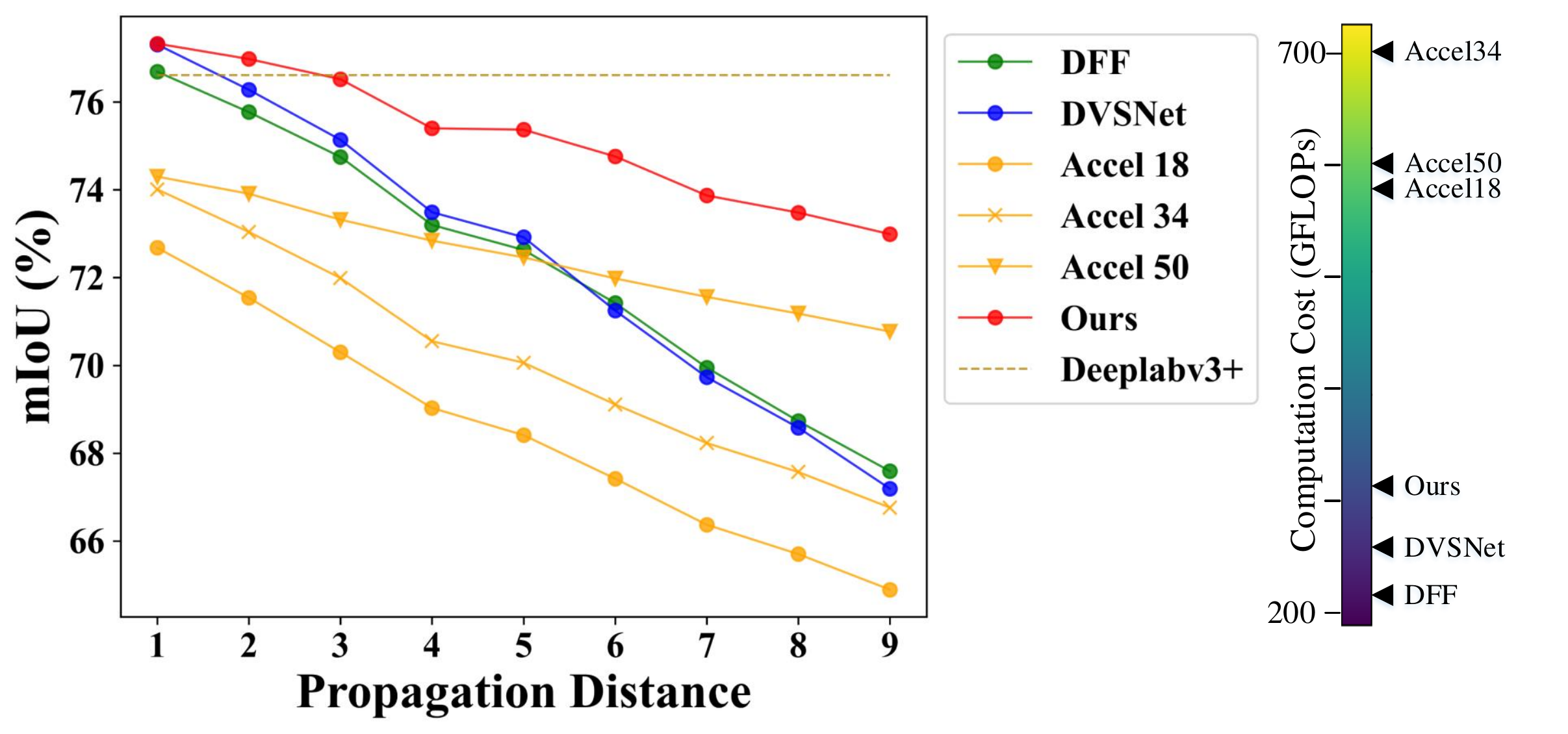}
	\end{center}
	\caption{\textbf{Performance evaluation on Cityscapes val subset with PDA Curve.} All methods are equipped with Deeplabv3+ as the backbone of segmentation network for fair comparison. The colorbar represents the computation cost of different methods for the propagation distance $D_P=5$, in which lighter color indicates higher computation cost. Best viewed in color.}
	\label{exp_SCM_cityscapes}
\end{figure}

\begin{figure}[t]
	\begin{center}
		\includegraphics[width=1.0\linewidth]{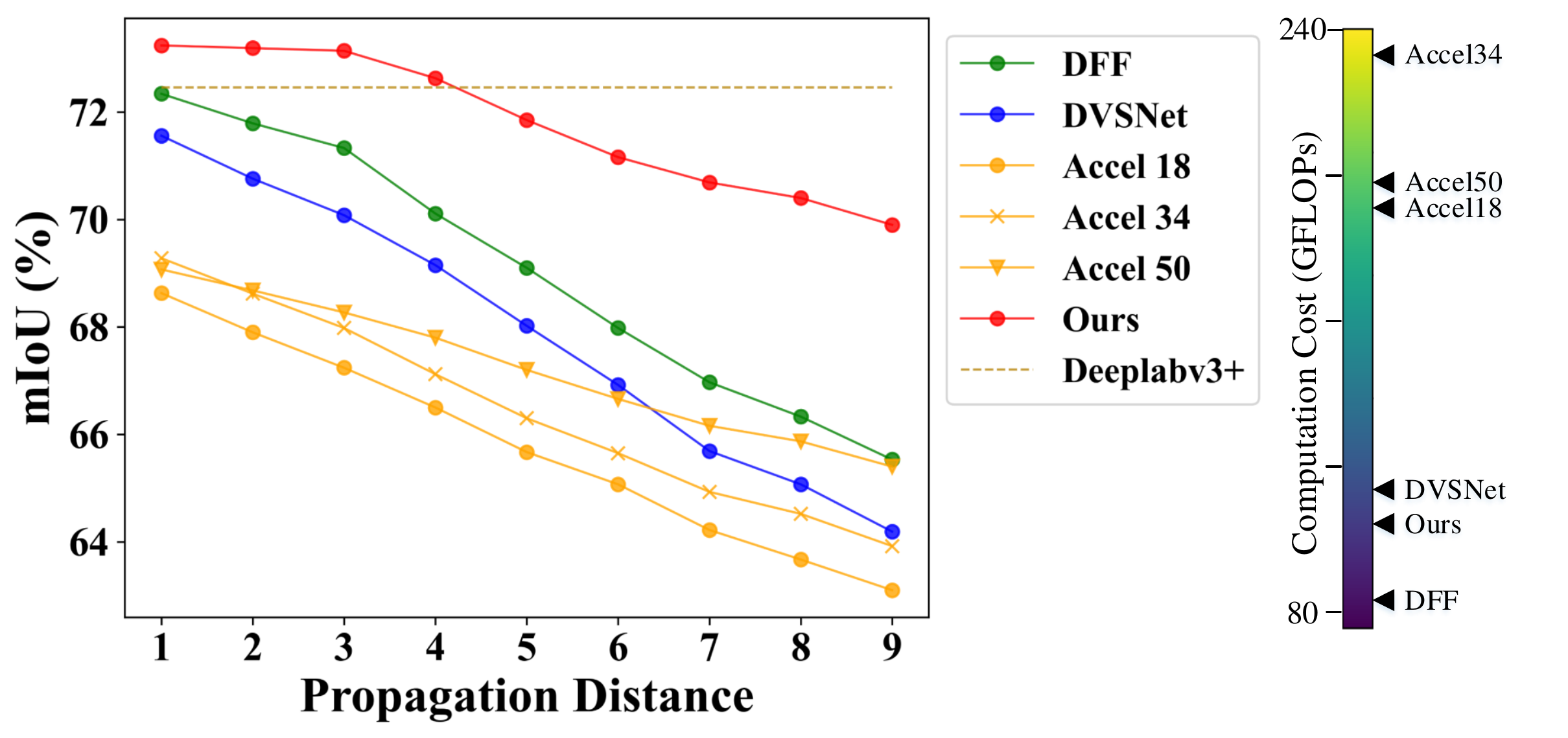}
	\end{center}
	\caption{\textbf{Performance evaluation on CamVid with PDA Curve.} All methods are equipped with Deeplabv3+ as the backbone of segmentation network for fair comparison. The colorbar represents the computation cost of different methods for the propagation distance $D_P=5$, in which lighter color indicates higher computation cost. Best viewed in color.}
	\label{exp_SCM_camvid}
\end{figure}

\begin{figure}[t]
	\begin{center}
		\includegraphics[width=1.0\linewidth]{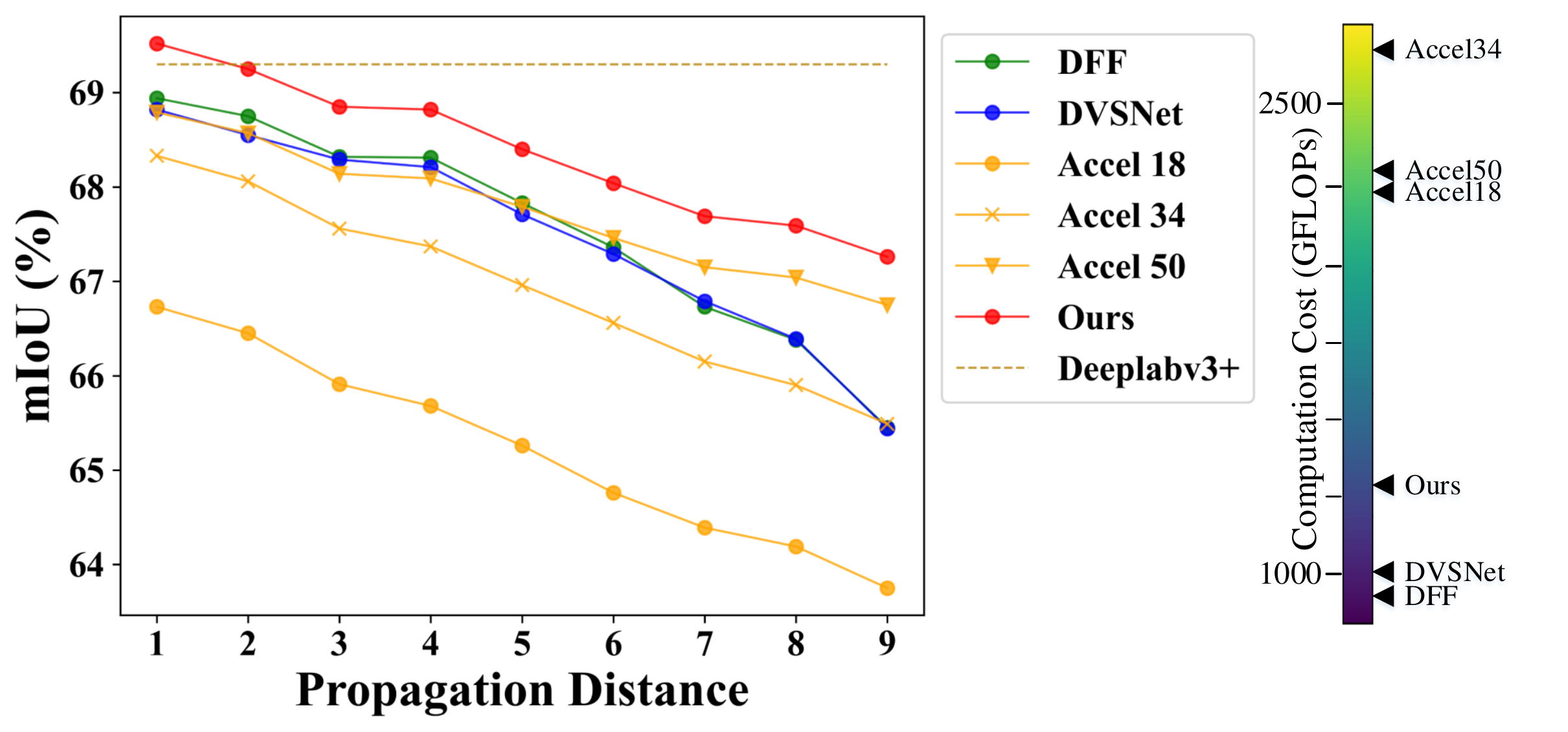}
	\end{center}
	\caption{\textbf{Performance evaluation on UAVid with PDA Curve.} All methods are equipped with Deeplabv3+ as the backbone of segmentation network for fair comparison. The colorbar represents the computation cost of different methods for the propagation distance $D_P=5$, in which lighter color indicates higher computation cost. Best viewed in color.}
	\label{exp_SCM_uavid}
\end{figure}

\subsection{Evaluation Metrics}
We experimentally evaluate different video semantic segmentation methods by measuring the segmentation accuracy and computational efficiency. 

For segmentation accuracy, we propose to use \emph{propagation distance \vs accuracy curve} (PDA Curve), which indicates how the segmentation accuracy changes along different propagation distances. Some previous works~\cite{zhu2017deep,jain2019accel} use the average accuracy among different propagation distances, which is inconvenient to figure out the actual performance. For computational efficiency, we propose to use \emph{computation cost \vs accuracy curve} (CCA Curve). CCA Curve is an important metric for model deployment, which indicates how the segmentation accuracy changes along different average computation cost.
Note that the PDA and CCA represent similar information on the segmentation performance since different computation costs are actually obtained by setting different propagation distances.

In the experiments on Cityscapes, we set the $11${th} to $19${th} frames as the key frame candidates, and propagate the feature of the selected key frame to the annotated $20${th} frame, which is used to measure the segmentation accuracy for each video clip. That is, the propagation distance (denoted by $D_{P}$) ranges from $1$ to $9$ for plotting the PDA Curve. When plotting the CCA Curve, we first calculate the computation cost of components used on the key frames (\ie, Net$_{seg}$) and non-key frames (\ie, FlowNet, CFNet, and DMNet), which are denoted by C$_{seg}$ and C$_{warp}$, respectively. The average computation cost is calculated by
\begin{equation}
C_{mean}=(C_{seg}+C_{warp}*D_{P})/(D_{P}+1). \label{eqn:cost}
\end{equation}
The evaluation on CamVid and UAVid is similar to Cityscapes. Here the mean intersection over union (mIoU) is adopted to measure the segmentation accuracy, and floating point operations (FLOPs) is used for the computation cost. Following common practices~\cite{molchanov2019pruning,howard2017mobilenets}, we calculate the FLOPs of convolutional layer, batch normalization layer, activation layer, and bilinear upsampling operator, of which the formulas are provided in Table~\ref{flops}.

\subsection{Performance Comparison}

We compare our proposed method with recent state-of-the-art methods, including DFF~\cite{zhu2017deep}, DVSNet~\cite{xu2018dynamic}, and Accel~\cite{jain2019accel}, and the CCA Curve is used for evaluation. 
Considering the baseline methods only provide the model on Cityscapes, here we only give the results on Cityscapes for fair comparison (the results on other datasets using our implementation will be presented in ablation study). To be specific, DFF and DVSNet use the same network DeeplabFast  as the segmentation backbone. But DVSNet splits the input frames into four overlapped regions to perform multiple times of segmentation, which is obviously more time-consuming. As for Accel, Deeplab with deformable ResNet-101 is used for image segmentation, and multiple versions of ResNets with different depths are adopted to process the current frame. Fig.~\ref{exp_cityscapes} shows the results of different methods on Cityscapes val subset. It can be seen that our proposed method significantly outperforms other method in both accuracy and efficiency. 

\subsection{Ablation Study}

\subsubsection{Effectiveness of our method}

Here we verify the effectiveness of our method on Cityscapes, CamVid, and UAVid, and the results are shown in Fig.~\ref{exp_SCM_cityscapes}, Fig.~\ref{exp_SCM_camvid}, and Fig.~\ref{exp_SCM_uavid}, respectively. For fair comparison, we reimplement the baseline methods with DeepLabv3+ as the backbone of segmentation networks and same FlowNet as in our proposed method. In particular, our implemented DeepLabv3+ achieves a mIoU score of $76.61\%$ on Cityscapes, $72.46\%$ on CamVid, and $69.30\%$ on UAVid for per-frame image segmentation. From the results, it can be seen that our proposed method significantly outperforms other state-of-the-art methods, especially for long-distance feature propagation. 

Besides, we calculate the average computation cost by fixing the propagation distance as $5$ for all methods. The results are shown in Fig.~\ref{exp_SCM_cityscapes}, Fig.~\ref{exp_SCM_camvid}, and Fig.~\ref{exp_SCM_uavid} with color bars, in which lighter color represents higher computation cost. Note that the computation cost of Accel34 is higher than that of Accel 50 because an extra deconvolutional layer is involved in Accel34 for feature upsampling. Each component in our proposed framework has a complexity of $O(HW)$, where $H$ and $W$ are height and width of the input images. Moreover, we analyze the computation cost of the main components and overall framework for key and non-key frames, and the statistics are provided in Table~\ref{computaion}. Note that the computation cost for key frames is that of the image segmentation model. It can be seen that the image segmentation network dominates the computation cost. As shown in Fig.~\ref{exp_SCM_cityscapes}, Fig.~\ref{exp_SCM_camvid}, and Fig.~\ref{exp_SCM_uavid}, our method has slightly higher computation cost than DFF and DVSNet, but gets significant accuracy improvement.

Actually, the key of the proposed method getting accuracy improvement and efficient computation is our designed distortion-aware mechanism. Benefited from such a design, both the feature extraction from current frames and feature correction can be completed by a lightweight network (e.g., DMNet and CFNet) due to only handling part of an image.  Moreover, only correcting features in distorted regions can effectively avoid false correction and further boost segmentation accuracy. Therefore, our method can outperform state-of-the-art methods in terms of accuracy and computation cost.

\begin{table}[t]
	\caption{Computation cost of different modules (GFLOPs). The resolution of input images is $1024\times2048$ on Cityscapes, $720\times960$ on CamVid, and $2160\times3840$ on UAVid.}
	\begin{center}
		\renewcommand{\arraystretch}{1.3}
		\begin{tabular}{l|c|c|c}
			\hline
			Module             	& Cityscapes	& CamVid	& UAVid    	\\
			\hline
			FlowNet 			& 86.918		& 29.219	& 341.441	\\
			CFNet 				& 123.775		& 41.741	& 484.319	\\
			DMNet 				& 0.400			& 0.132		& 1.584		\\
			\hline
			Overall for key		& 826.378		& 272.189	& 3265.457	\\
			Overall for non-key	& 212.910		& 71.448	& 830.515	\\
			\hline
		\end{tabular}
	\end{center}
	\label{computaion}
\end{table}

It is notable that our method can yield higher segmentation accuracy than per-frame image segmentation for short-distance feature propagation. It is because our proposed feature propagation can well exploit the information from multiple frames. That is, the segmentation of the current frame would benefit from the feature combination of the previous and current frames. 

\subsubsection{Effect of different components}

\begin{table}[t]
	\caption{Effect of different components in our method on Cityscapes val subset. ''mIoU'' is used as the metric.}
	\begin{center}
		\renewcommand{\arraystretch}{1.3}
		\begin{tabular}{c|c|c|c|c|c|c}
			\hline
			\multirow{2}{*}{DDS} & \multicolumn{2}{c|}{FCM} & \multicolumn{4}{c}{Distance} 							\\
			\cline{2-7}
			& DGFL		& DGFC      & 1			& 5				& 9 			& Mean   	 						\\
			\hline
			$\surd$ 	& $\surd$	& $\surd$	& \bf{77.33}	& \bf{75.37}	& \bf{72.99}	& \bf{75.19} 		\\
			& $\surd$	& $\surd$	& 77.30		& 74.30			& 71.37			& 74.26	(-0.93)						\\
			$\surd$ 	& 			& $\surd$	& 77.09			& 74.34			& 71.16			& 74.17	(-1.02)		\\
			$\surd$ 	& $\surd$	& 			& 74.98			& 73.65			& 71.92			& 73.56	(-1.63)		\\
			$\surd$ 	& 			& 			& 76.64			& 72.45 		& 67.53 		& 72.21	(-2.98)		\\
			\hline
		\end{tabular}
	\end{center}
	\label{exp_distortion}
\end{table}

\begin{table}[t]
	\caption{Upper bound analysis of different methods on Cityscapes val subset. mIoU is used as the metric. * denotes the upper bound, and $\uparrow$ represents the corresponding gap.}
	\begin{center}
		\renewcommand{\arraystretch}{1.3}
		\begin{tabular}{l|c|c|c|c|c}
			\hline
			Distance            & 1				& 5				& 9 			& Mean   		& $\uparrow$   	\\
			\hline
			DFF          		& 73.94			& 69.06 		& 64.04 		& 69.14			&				\\
			DFF*	 			& 75.47			& 71.16			& 66.31			& 71.15 		& 2.01			\\
			\hline
			Accel18 			& 73.19			& 67.59			& 64.54			& 68.21			&				\\
			Accel18* 			& 77.37			& 70.44			& 66.92			& 71.26 		& 3.05			\\
			\hline
			Accel34 			& 73.01			& 69.03			& 67.19			& 69.64			&				\\
			Accel34* 			& 78.33			& 73.08			& 70.76			& 73.82			& 4.18			\\
			\hline
			Accel50 			& 74.55			& 72.48			& 71.27			& 72.70			&				\\
			Accel50* 			& 78.97			& 75.98			& 74.52			& 76.36			& 3.66			\\
			\hline
			Ours 				& \bf{77.33}	& \bf{75.37}	& \bf{72.99}	& \bf{75.19}	&				\\
			Ours*	 			& \bf{77.99}	& \bf{76.59}	& \bf{74.50}	& \bf{76.37}	& \bf{1.18}		\\
			\hline
		\end{tabular}
	\end{center}
	\label{exp_upper_bound}
\end{table}

Here we investigate the contribution of each proposed component to the segmentation performance by removing them (\ie, DDS, DGFL, and DGFC) one by one. Table~\ref{exp_distortion} gives the results, in which the propagation distances $\{1,5,9\}$ are particularly used and the mean segmentation accuracy over all distances is also provided.
From the results, we have the following observations. (1) DDS can improve the network training of our proposed method and further bring accuracy increase, as shown in the first two rows.  (2) FCM is the main source of performance gains, especially for long-distance feature propagation that usually would cause serious distortion. (3) DGFL and DGFC in FCM are both important. In particular, CFNet without DGFL cannot effectively extract the correction cues, and without DGFC, the false correction would become severe, especially for short-distance propagation. 
Considering these results, it is convinced that our proposed distortion-aware feature correction is very effective for boosting the performance of propagation-based video semantic segmentation.

\subsubsection{Upper bound analysis}

Similar to our proposed method, DFF~\cite{zhu2017deep} and Accel~\cite{zhu2017deep} also rectify the propagated features. Here we explore the upper bound of segmentation accuracy of these methods, in which only the wrongly predicted regions are rectified during inference (the ground truth of semantic segmentation is used). 
Table~\ref{exp_upper_bound} shows the results. It can be seen that DFF and Accel have a larger gap corresponding to their upper bounds while our method can achieve a smaller one. Such results imply that our method can effectively alleviate false correction with our proposed distortion prediction.

\subsubsection{Design of Distortion Map Prediction}

\begin{figure}[t]
	\begin{center}
		\includegraphics[width=1.0\linewidth]{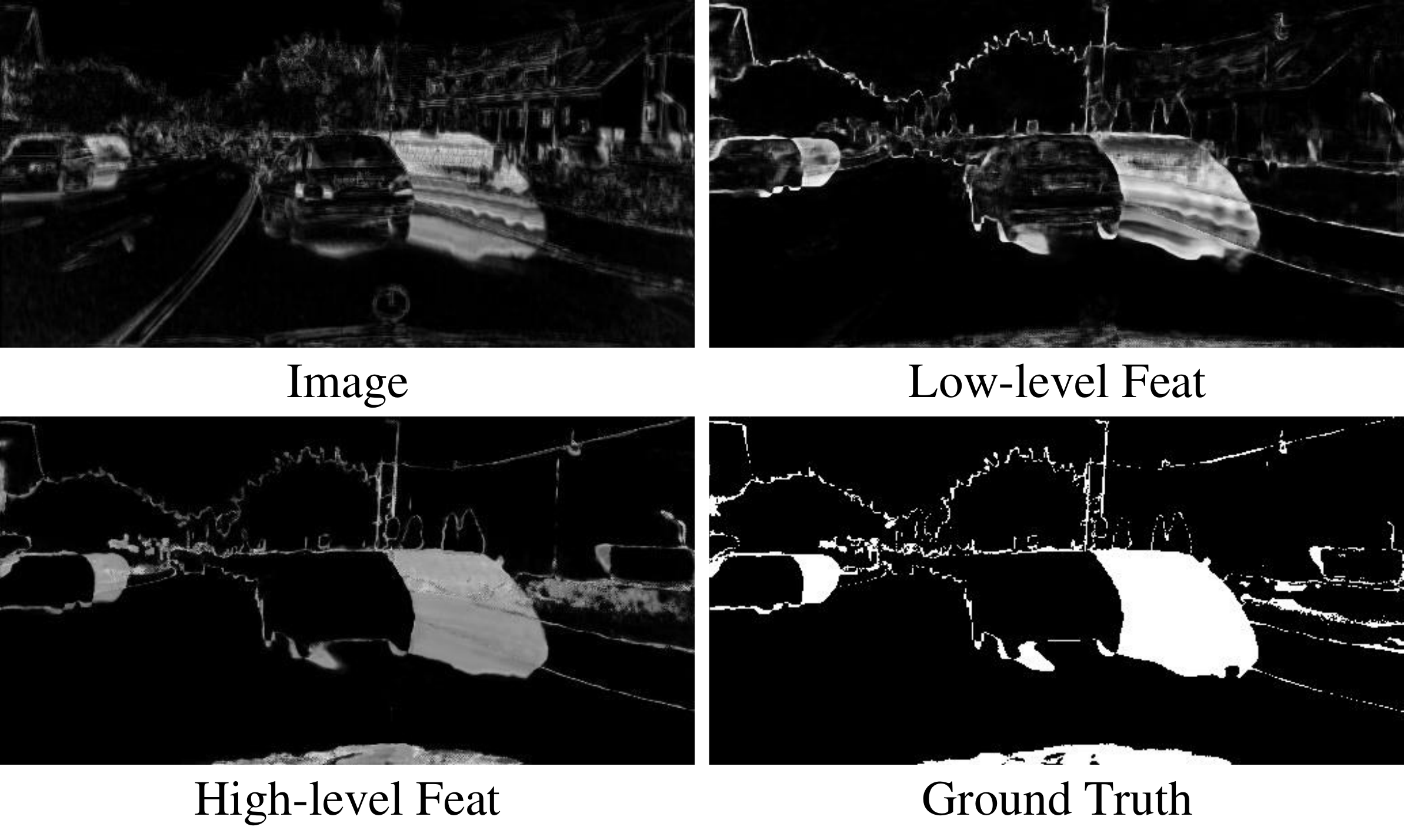}
	\end{center}
	\caption{\textbf{The predicted distortion maps using different features.} It can be seen that the  higher-level  feature can get better distortion maps.}
	\label{distortion_input}
\end{figure}

In this work, we propose to predict the distortion maps using images rather than features in order to achieve low computation cost. Here we compare different data to predict distortion maps regardless of the computational price. In particular, we take the propagated features after classifier (high-level feature) and features after entry flow block2 in DeepLabv3+ (low-level feature) as the inputs of DMNet. Besides, we use the ground truth of distortion maps to test the upper bound of segmentation performance, which can well demonstrate the effectiveness of our idea to exploit the distortion map in video semantic segmentation. 

\begin{figure}[t]
	\begin{center}
		\includegraphics[width=1.0\linewidth]{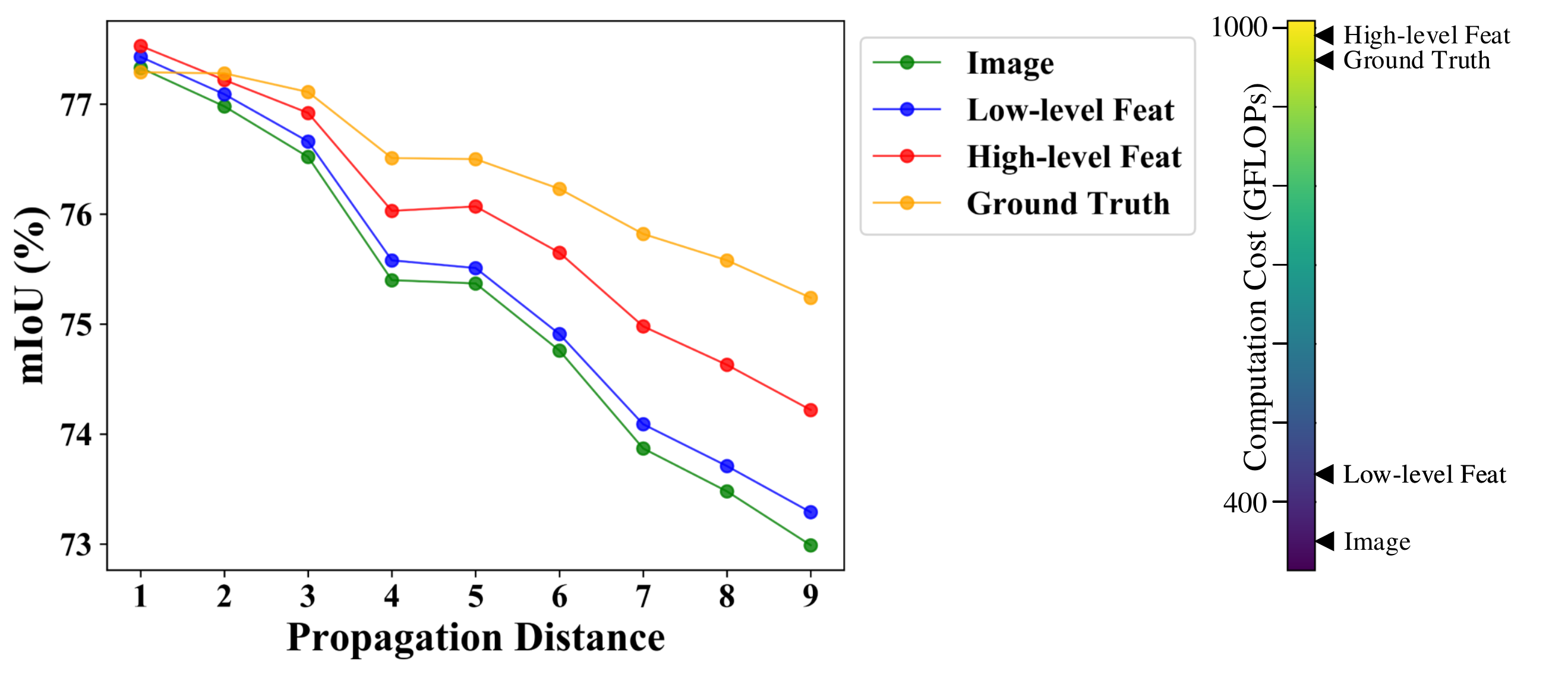}
	\end{center}
	\caption{\textbf{Comparison of different data to predict distortion maps.} Here the segmentation accuracy on Cityscapes val subset is adopted for evaluation. It can be seen that higher-level feature can generate higher-quality distortion maps but would involve higher computation cost. Best viewed in color.}	
	\label{exp_distortion_input_with_computation}
\end{figure}

Fig.~\ref{distortion_input} provides the visual comparison of predicted distortion maps for different features, and Fig.~\ref{exp_distortion_input_with_computation} gives the corresponding segmentation performance. We have the following observations. First, the higher-level feature can bring higher segmentation accuracy for getting more consistent distortion maps with the ground truth, but would involve higher computation cost. Thus we need to find a good trade-off between the segmentation accuracy and computation cost. Second, from the results in Table~\ref{exp_distortion} and Fig.~\ref{exp_distortion_input_with_computation}, it can be seen that our proposed method can get significant performance improvement if the ground truths of distortion maps are used, which shows the rationality of focusing on the distortion regions in this work.

\subsubsection{Visualization}

\begin{figure*}[t]
	\begin{center}
		\includegraphics[width=1.0\linewidth]{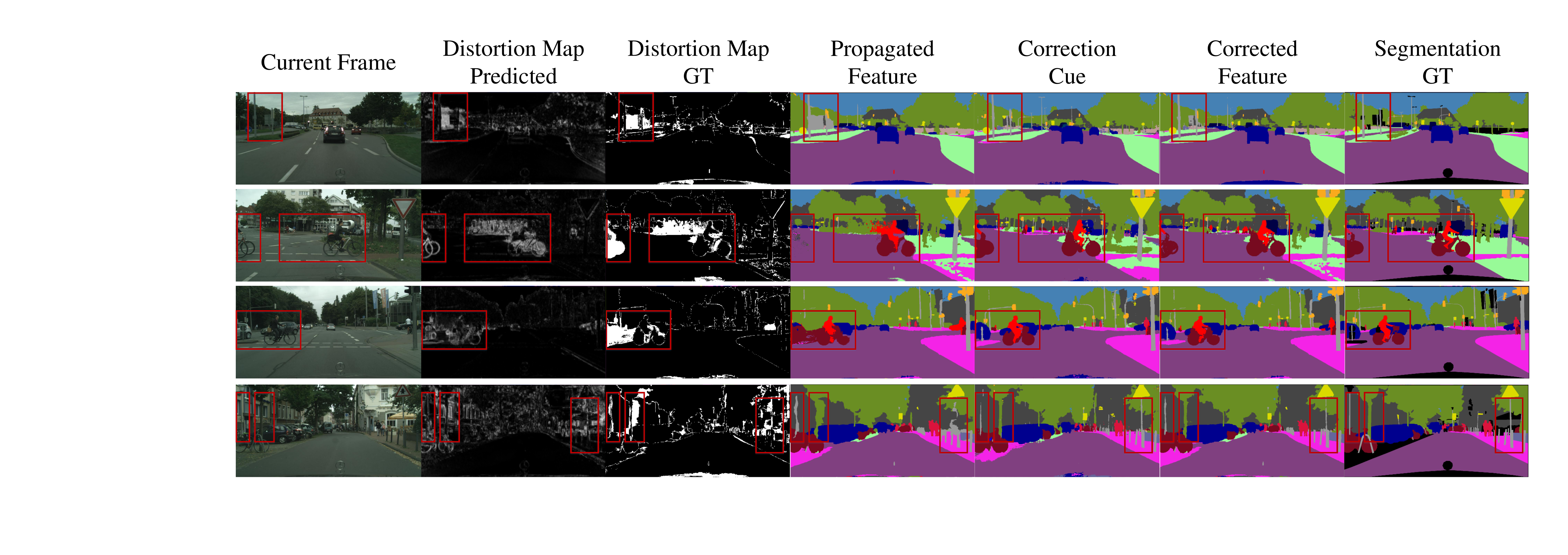}
	\end{center}
	\caption{\textbf{Visualization of some samples from Cityscapes.} It can be seen that the predicted distortion map can represent the distortion pattern of propagated features, and our proposed method can effectively correct the distorted features. Red rectangles highlight the main distorted regions. Best viewed in color.}	
	\label{visualization}
\end{figure*}

To intuitively illustrate our proposed method, we provide the visualizations of four samples from the Cityscapes datasets in Fig.~\ref{visualization}, where the intermediate features are demonstrated by applying the segmentation head to them. For each sample, we extract the feature from the key frame, and then propagate it to the current frame (the $9$th one from the key frame). First, we can see that the predicted distortion maps can represent the distorted regions of propagated features. Second, under the guidance of distortion maps, our proposed method can accurately extract the correction cues, especially on the distorted regions, and then effectively correct the propagated features.

\section{Conclusion} \label{sec:conclusion}
We present a novel video semantic segmentation method in this paper, aiming at achieving high segmentation accuracy and competitive real-time performance simultaneously by tackling the feature distortion problem in propagation. Specifically, we propose DMNet to predict distorted regions of the propagated features, and then propose FCM to correct the distorted features with a lightweight model. Our experimental results on Cityscapes, CamVid, and UAVid show that the proposed method outperforms the state-of-the-art methods in both precision and speed.

\section*{Acknowledge} \label{sec:acknowledge}
We acknowledge the support of GPU cluster built by MCC Lab of Information Science and Technology Institution, USTC.


%





\ifCLASSOPTIONcaptionsoff
\newpage
\fi



%

\bibliographystyle{IEEEtran}

\begin{thebibliography}{10}
	\providecommand{\url}[1]{#1}
	\csname url@samestyle\endcsname
	\providecommand{\newblock}{\relax}
	\providecommand{\bibinfo}[2]{#2}
	\providecommand{\BIBentrySTDinterwordspacing}{\spaceskip=0pt\relax}
	\providecommand{\BIBentryALTinterwordstretchfactor}{4}
	\providecommand{\BIBentryALTinterwordspacing}{\spaceskip=\fontdimen2\font plus
		\BIBentryALTinterwordstretchfactor\fontdimen3\font minus
		\fontdimen4\font\relax}
	\providecommand{\BIBforeignlanguage}[2]{{%
			\expandafter\ifx\csname l@#1\endcsname\relax
			\typeout{** WARNING: IEEEtran.bst: No hyphenation pattern has been}%
			\typeout{** loaded for the language `#1'. Using the pattern for}%
			\typeout{** the default language instead.}%
			\else
			\language=\csname l@#1\endcsname
			\fi
			#2}}
	\providecommand{\BIBdecl}{\relax}
	\BIBdecl
	
	\bibitem{long2015fully}
	J.~Long, E.~Shelhamer, and T.~Darrell, ``Fully convolutional networks for
	semantic segmentation,'' in \emph{CVPR}, 2015.
	
	\bibitem{Cordts_2016_CVPR}
	M.~Cordts, M.~Omran, S.~Ramos, T.~Rehfeld, M.~Enzweiler, R.~Benenson,
	U.~Franke, S.~Roth, and B.~Schiele, ``The cityscapes dataset for semantic
	urban scene understanding,'' in \emph{CVPR}, 2016.
	
	\bibitem{brostow2009semantic}
	G.~J. Brostow, J.~Fauqueur, and R.~Cipolla, ``Semantic object classes in video:
	A high-definition ground truth database,'' \emph{Pattern Recognition
		Letters}, 2009.
	
	\bibitem{lyu2020uavid}
	Y.~Lyu, G.~Vosselman, G.-S. Xia, A.~Yilmaz, and M.~Y. Yang, ``Uavid: A semantic
	segmentation dataset for uav imagery,'' \emph{ISPRS Journal of Photogrammetry
		and Remote Sensing}, 2020.
	
	\bibitem{kostavelis2015semantic}
	I.~Kostavelis and A.~Gasteratos, ``Semantic mapping for mobile robotics tasks:
	A survey,'' \emph{Robotics and Autonomous Systems}, 2015.
	
	\bibitem{teichmann2018multinet}
	M.~Teichmann, M.~Weber, M.~Zoellner, R.~Cipolla, and R.~Urtasun, ``Multinet:
	Real-time joint semantic reasoning for autonomous driving,'' in \emph{IV},
	2018.
	
	\bibitem{Liu_2017_CVPR}
	S.~Liu, C.~Wang, R.~Qian, H.~Yu, R.~Bao, and Y.~Sun, ``Surveillance video
	parsing with single frame supervision,'' in \emph{CVPR}, 2017.
	
	\bibitem{zhu2017deep}
	X.~Zhu, Y.~Xiong, J.~Dai, L.~Yuan, and Y.~Wei, ``Deep feature flow for video
	recognition,'' in \emph{CVPR}, 2017.
	
	\bibitem{xu2018dynamic}
	Y.-S. Xu, T.-J. Fu, H.-K. Yang, and C.-Y. Lee, ``Dynamic video segmentation
	network,'' in \emph{CVPR}, 2018.
	
	\bibitem{gadde2017semantic}
	R.~Gadde, V.~Jampani, and P.~V. Gehler, ``Semantic video cnns through
	representation warping,'' in \emph{ICCV}, 2017.
	
	\bibitem{jain2019accel}
	S.~Jain, X.~Wang, and J.~E. Gonzalez, ``Accel: A corrective fusion network for
	efficient semantic segmentation on video,'' in \emph{CVPR}, 2019.
	
	\bibitem{dosovitskiy2015flownet}
	A.~Dosovitskiy, P.~Fischer, E.~Ilg, P.~Hausser, C.~Hazirbas, V.~Golkov, P.~Van
	Der~Smagt, D.~Cremers, and T.~Brox, ``Flownet: Learning optical flow with
	convolutional networks,'' in \emph{ICCV}, 2015.
	
	\bibitem{ilg2017flownet}
	E.~Ilg, N.~Mayer, T.~Saikia, M.~Keuper, A.~Dosovitskiy, and T.~Brox, ``Flownet
	2.0: Evolution of optical flow estimation with deep networks,'' in
	\emph{CVPR}, 2017.
	
	\bibitem{liu2019selflow}
	P.~Liu, M.~Lyu, I.~King, and J.~Xu, ``Selflow: Self-supervised learning of
	optical flow,'' in \emph{CVPR}, 2019.
	
	\bibitem{neoral2018continual}
	M.~Neoral, J.~{\v{S}}ochman, and J.~Matas, ``Continual occlusion and optical
	flow estimation,'' in \emph{ACCV}, 2018.
	
	\bibitem{iandola2016squeezenet}
	F.~N. Iandola, S.~Han, M.~W. Moskewicz, K.~Ashraf, W.~J. Dally, and K.~Keutzer,
	``Squeezenet: Alexnet-level accuracy with 50x fewer parameters and< 0.5 mb
	model size,'' \emph{arXiv preprint arXiv:1602.07360}, 2016.
	
	\bibitem{simonyan2014very}
	K.~Simonyan and A.~Zisserman, ``Very deep convolutional networks for
	large-scale image recognition,'' \emph{arXiv preprint arXiv:1409.1556}, 2014.
	
	\bibitem{he2016deep}
	K.~He, X.~Zhang, S.~Ren, and J.~Sun, ``Deep residual learning for image
	recognition,'' in \emph{CVPR}, 2016.
	
	\bibitem{szegedy2015going}
	C.~Szegedy, W.~Liu, Y.~Jia, P.~Sermanet, S.~Reed, D.~Anguelov, D.~Erhan,
	V.~Vanhoucke, and A.~Rabinovich, ``Going deeper with convolutions,'' in
	\emph{CVPR}, 2015.
	
	\bibitem{huang2017densely}
	G.~Huang, Z.~Liu, L.~Van Der~Maaten, and K.~Q. Weinberger, ``Densely connected
	convolutional networks,'' in \emph{CVPR}, 2017.
	
	\bibitem{zhao2017pyramid}
	H.~Zhao, J.~Shi, X.~Qi, X.~Wang, and J.~Jia, ``Pyramid scene parsing network,''
	in \emph{CVPR}, 2017.
	
	\bibitem{wu2019wider}
	Z.~Wu, C.~Shen, and A.~Van Den~Hengel, ``Wider or deeper: Revisiting the resnet
	model for visual recognition,'' \emph{Pattern Recognition}, 2019.
	
	\bibitem{lin2017refinenet}
	G.~Lin, A.~Milan, C.~Shen, and I.~Reid, ``Refinenet: Multi-path refinement
	networks for high-resolution semantic segmentation,'' in \emph{CVPR}, 2017.
	
	\bibitem{chen2018deeplab}
	L.-C. Chen, G.~Papandreou, I.~Kokkinos, K.~Murphy, and A.~L. Yuille, ``Deeplab:
	Semantic image segmentation with deep convolutional nets, atrous convolution,
	and fully connected crfs,'' \emph{TPAMI}, 2018.
	
	\bibitem{yu2015multi}
	F.~Yu and V.~Koltun, ``Multi-scale context aggregation by dilated
	convolutions,'' \emph{arXiv preprint arXiv:1511.07122}, 2015.
	
	\bibitem{chen2014semantic}
	L.-C. Chen, G.~Papandreou, I.~Kokkinos, K.~Murphy, and A.~L. Yuille, ``Semantic
	image segmentation with deep convolutional nets and fully connected crfs,''
	\emph{arXiv preprint arXiv:1412.7062}, 2014.
	
	\bibitem{zheng2015conditional}
	S.~Zheng, S.~Jayasumana, B.~Romera-Paredes, V.~Vineet, Z.~Su, D.~Du, C.~Huang,
	and P.~H. Torr, ``Conditional random fields as recurrent neural networks,''
	in \emph{ICCV}, 2015.
	
	\bibitem{he2015spatial}
	K.~He, X.~Zhang, S.~Ren, and J.~Sun, ``Spatial pyramid pooling in deep
	convolutional networks for visual recognition,'' \emph{TPAMI}, 2015.
	
	\bibitem{chen2017rethinking}
	L.-C. Chen, G.~Papandreou, F.~Schroff, and H.~Adam, ``Rethinking atrous
	convolution for semantic image segmentation,'' \emph{arXiv preprint
		arXiv:1706.05587}, 2017.
	
	\bibitem{wang2018detecting}
	Q.~Wang, M.~Chen, F.~Nie, and X.~Li, ``Detecting coherent groups in crowd
	scenes by multiview clustering,'' \emph{TPAMI}, 2018.
	
	\bibitem{wang2017joint}
	Q.~Wang, J.~Gao, and Y.~Yuan, ``A joint convolutional neural networks and
	context transfer for street scenes labeling,'' \emph{TITS}, 2017.
	
	\bibitem{Huang_2019_ICCV}
	Z.~Huang, X.~Wang, L.~Huang, C.~Huang, Y.~Wei, and W.~Liu, ``Ccnet: Criss-cross
	attention for semantic segmentation,'' in \emph{ICCV}, 2019.
	
	\bibitem{Sun_2019_CVPR}
	K.~Sun, B.~Xiao, D.~Liu, and J.~Wang, ``Deep high-resolution representation
	learning for human pose estimation,'' in \emph{CVPR}, 2019.
	
	\bibitem{Zhong_2020_CVPR}
	Z.~Zhong, Z.~Q. Lin, R.~Bidart, X.~Hu, I.~B. Daya, Z.~Li, W.-S. Zheng, J.~Li,
	and A.~Wong, ``Squeeze-and-attention networks for semantic segmentation,'' in
	\emph{CVPR}, 2020.
	
	\bibitem{Li_2020_CVPR}
	Y.~Li, L.~Song, Y.~Chen, Z.~Li, X.~Zhang, X.~Wang, and J.~Sun, ``Learning
	dynamic routing for semantic segmentation,'' in \emph{CVPR}, 2020.
	
	\bibitem{lin2020cross}
	K.~Lin, L.~Wang, K.~Luo, Y.~Chen, Z.~Liu, and M.-T. Sun, ``Cross-domain
	complementary learning using pose for multi-person part segmentation,''
	\emph{TCSVT}, 2020.
	
	\bibitem{ji2020encoder}
	J.~Ji, R.~Shi, S.~Li, P.~Chen, and Q.~Miao, ``Encoder-decoder with cascaded
	crfs for semantic segmentation,'' \emph{TCSVT}, 2020.
	
	\bibitem{li2018low}
	Y.~Li, J.~Shi, and D.~Lin, ``Low-latency video semantic segmentation,'' in
	\emph{CVPR}, 2018.
	
	\bibitem{horn1981determining}
	B.~K. Horn and B.~G. Schunck, ``Determining optical flow,'' \emph{Artificial
		intelligence}, 1981.
	
	\bibitem{anguita2009optimization}
	M.~Anguita, J.~D{\'\i}az, E.~Ros, and F.~J. Fern{\'a}ndez-Baldomero,
	``Optimization strategies for high-performance computing of optical-flow in
	general-purpose processors,'' \emph{TCSVT}, 2009.
	
	\bibitem{sun2018pwc}
	D.~Sun, X.~Yang, M.-Y. Liu, and J.~Kautz, ``Pwc-net: Cnns for optical flow
	using pyramid, warping, and cost volume,'' in \emph{CVPR}, 2018.
	
	\bibitem{zhai2019optical}
	M.~Zhai, X.~Xiang, R.~Zhang, N.~Lv, and A.~El~Saddik, ``Optical flow estimation
	using dual self-attention pyramid networks,'' \emph{TCSVT}, 2019.
	
	\bibitem{chen2016full}
	Q.~Chen and V.~Koltun, ``Full flow: Optical flow estimation by global
	optimization over regular grids,'' in \emph{CVPR}, 2016.
	
	\bibitem{sundaram2010dense}
	N.~Sundaram, T.~Brox, and K.~Keutzer, ``Dense point trajectories by
	gpu-accelerated large displacement optical flow,'' in \emph{ECCV}, 2010.
	
	\bibitem{ranjan2017optical}
	A.~Ranjan and M.~J. Black, ``Optical flow estimation using a spatial pyramid
	network,'' in \emph{CVPR}, 2017.
	
	\bibitem{hu2018squeeze}
	J.~Hu, L.~Shen, and G.~Sun, ``Squeeze-and-excitation networks,'' in
	\emph{CVPR}, 2018.
	
	\bibitem{yuan2018ocnet}
	Y.~Yuan and J.~Wang, ``Ocnet: Object context network for scene parsing,''
	\emph{arXiv preprint arXiv:1809.00916}, 2018.
	
	\bibitem{fu2019dual}
	J.~Fu, J.~Liu, H.~Tian, Y.~Li, Y.~Bao, Z.~Fang, and H.~Lu, ``Dual attention
	network for scene segmentation,'' in \emph{CVPR}, 2019.
	
	\bibitem{zhao2018psanet}
	H.~Zhao, Y.~Zhang, S.~Liu, J.~Shi, C.~Change~Loy, D.~Lin, and J.~Jia, ``Psanet:
	Point-wise spatial attention network for scene parsing,'' in \emph{ECCV},
	2018.
	
	\bibitem{chen2018reverse}
	S.~Chen, X.~Tan, B.~Wang, and X.~Hu, ``Reverse attention for salient object
	detection,'' in \emph{ECCV}, 2018.
	
	\bibitem{chen2018embedding}
	S.~Chen, B.~Wang, X.~Tan, and X.~Hu, ``Embedding attention and residual network
	for accurate salient object detection,'' \emph{IEEE Transactions on
		Cybernetics}, 2018.
	
	\bibitem{zhang2020bilateral}
	Z.~Zhang, Z.~Lin, J.~Xu, W.~Jin, S.-P. Lu, and D.-P. Fan, ``Bilateral attention
	network for rgb-d salient object detection,'' \emph{arXiv preprint
		arXiv:2004.14582}, 2020.
	
	\bibitem{fan2020pranet}
	D.-P. Fan, G.-P. Ji, T.~Zhou, G.~Chen, H.~Fu, J.~Shen, and L.~Shao, ``Pranet:
	Parallel reverse attention network for polyp segmentation,'' in
	\emph{MICCAI}, 2020.
	
	\bibitem{fayyaz2016stfcn}
	M.~Fayyaz, M.~H. Saffar, M.~Sabokrou, M.~Fathy, F.~Huang, and R.~Klette,
	``Stfcn: spatio-temporal fully convolutional neural network for semantic
	segmentation of street scenes,'' in \emph{ACCV}, 2016.
	
	\bibitem{nilsson2018semantic}
	D.~Nilsson and C.~Sminchisescu, ``Semantic video segmentation by gated
	recurrent flow propagation,'' in \emph{CVPR}, 2018.
	
	\bibitem{tran2016deep}
	D.~Tran, L.~Bourdev, R.~Fergus, L.~Torresani, and M.~Paluri, ``Deep end2end
	voxel2voxel prediction,'' in \emph{CVPR workshops}, 2016.
	
	\bibitem{wang2019superpixel}
	Y.~Wang, W.~Ding, B.~Zhang, H.~Li, and S.~Liu, ``Superpixel labeling priors and
	mrf for aerial video segmentation,'' \emph{TCSVT}, 2019.
	
	\bibitem{shelhamer2016clockwork}
	E.~Shelhamer, K.~Rakelly, J.~Hoffman, and T.~Darrell, ``Clockwork convnets for
	video semantic segmentation,'' in \emph{ECCV}, 2016.
	
	\bibitem{hu2020temporally}
	P.~Hu, F.~Caba~Heilbron, O.~Wang, Z.~Lin, S.~Sclaroff, and F.~Perazzi,
	``Temporally distributed networks for fast video semantic segmentation,'' in
	\emph{CVPR}, 2020.
	
	\bibitem{wang2005interactive}
	J.~Wang, P.~Bhat, R.~A. Colburn, M.~Agrawala, and M.~F. Cohen, ``Interactive
	video cutout,'' \emph{ToG}, 2005.
	
	\bibitem{li2005video}
	Y.~Li, J.~Sun, and H.-Y. Shum, ``Video object cut and paste,'' in \emph{ACM
		SIGGRAPH}, 2005.
	
	\bibitem{price2009livecut}
	B.~L. Price, B.~S. Morse, and S.~Cohen, ``Livecut: Learning-based interactive
	video segmentation by evaluation of multiple propagated cues,'' in
	\emph{ICCV}, 2009.
	
	\bibitem{chen2019motion}
	Z.~Chen, C.~Guo, J.~Lai, and X.~Xie, ``Motion-appearance interactive encoding
	for object segmentation in unconstrained videos,'' \emph{TCSVT}, 2019.
	
	\bibitem{liu2020guided}
	W.~Liu, G.~Lin, T.~Zhang, and Z.~Liu, ``Guided co-segmentation network for fast
	video object segmentation,'' \emph{TCSVT}, 2020.
	
	\bibitem{chen2018encoder}
	L.-C. Chen, Y.~Zhu, G.~Papandreou, F.~Schroff, and H.~Adam, ``Encoder-decoder
	with atrous separable convolution for semantic image segmentation,'' in
	\emph{ECCV}, 2018.
	
	\bibitem{zou2018unsupervised}
	Y.~Zou, Z.~Yu, B.~Vijaya~Kumar, and J.~Wang, ``Unsupervised domain adaptation
	for semantic segmentation via class-balanced self-training,'' in \emph{ECCV},
	2018.
	
	\bibitem{hungadversarial}
	W.-C. Hung, Y.-H. Tsai, Y.-T. Liou34, Y.-Y. Lin, and M.-H. Yang15,
	``Adversarial learning for semi-supervised semantic segmentation,'' in
	\emph{BMVC}, 2018.
	
	\bibitem{kingma2014adam}
	D.~P. Kingma and J.~Ba, ``Adam: A method for stochastic optimization,'' in
	\emph{ICLR}, 2015.
	
	\bibitem{cordts2016cityscapes}
	M.~Cordts, M.~Omran, S.~Ramos, T.~Rehfeld, M.~Enzweiler, R.~Benenson,
	U.~Franke, S.~Roth, and B.~Schiele, ``The cityscapes dataset for semantic
	urban scene understanding,'' in \emph{CVPR}, 2016.
	
	\bibitem{molchanov2019pruning}
	P.~Molchanov, S.~Tyree, T.~Karras, T.~Aila, and J.~Kautz, ``Pruning
	convolutional neural networks for resource efficient inference,'' in
	\emph{ICLR}, 2019.
	
	\bibitem{howard2017mobilenets}
	A.~G. Howard, M.~Zhu, B.~Chen, D.~Kalenichenko, W.~Wang, T.~Weyand,
	M.~Andreetto, and H.~Adam, ``Mobilenets: Efficient convolutional neural
	networks for mobile vision applications,'' \emph{arXiv preprint
		arXiv:1704.04861}, 2017.
	
\end{thebibliography}

%

\newpage

\begin{IEEEbiography}[{\includegraphics[width=1in,height=1.25in,clip,keepaspectratio]{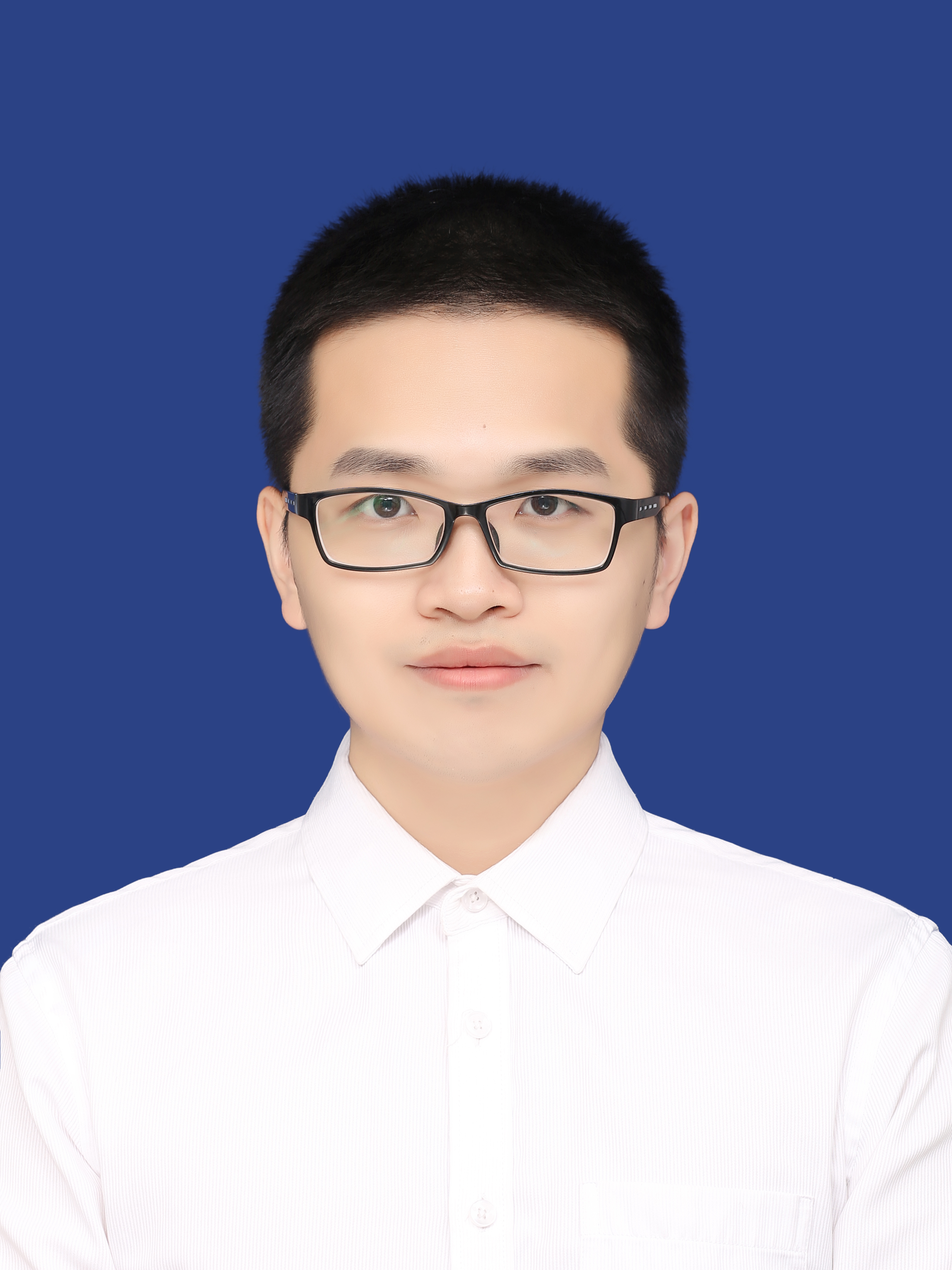}}]
	{Jiafan Zhuang}	received the B.S. degree in control science and engineering from the University of Science and Technology of China (USTC), Hefei, China, in 2017, and is currently working toward the Ph.D. degree in control science and engineering at the USTC.
	
	His current research interests include semantic segmentation and video analysis.
\end{IEEEbiography}

\vspace{-120mm}

\begin{IEEEbiography}[{\includegraphics[width=1in,height=1.25in,clip,keepaspectratio]{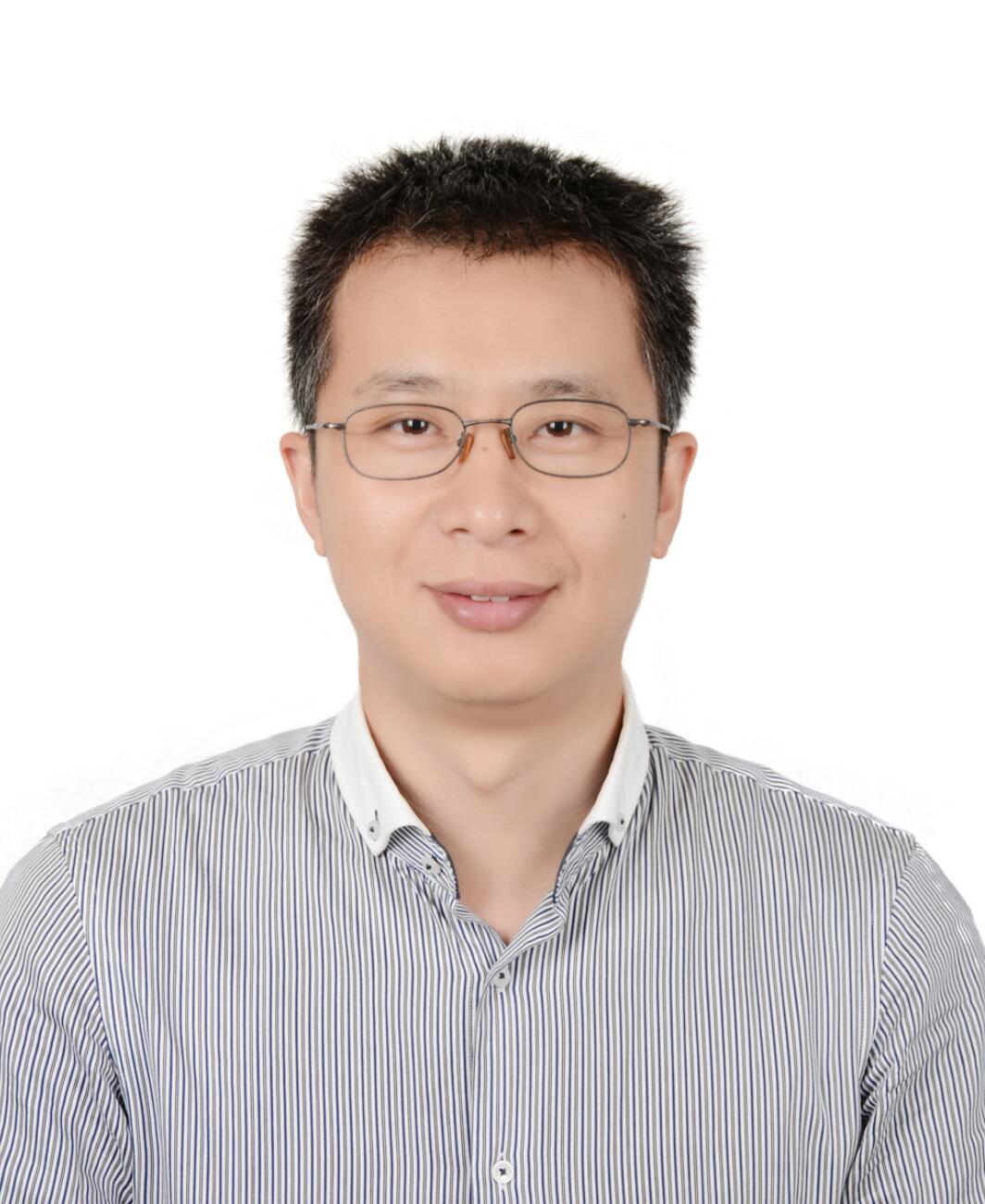}}]
	{Zilei Wang (M'13)}	received the B.S. and Ph.D.	degrees in control science and engineering from	the University of Science and Technology of China (USTC), Hefei, China, in 2002 and 2007, respectively.
	
	He is currently an Associate Professor with the	Department of Automation, USTC, and the Founding Lead of the Vision and Multimedia Research	Group (http://vim.ustc.edu.cn). His research interests include computer vision, multimedia, and deep learning.
	
	Prof. Wang is a Member of the Youth Innovation Promotion Association, Chinese Academy of Sciences.
\end{IEEEbiography}

\vspace{-120mm}

\begin{IEEEbiography}[{\includegraphics[width=1in,height=1.25in,clip,keepaspectratio]{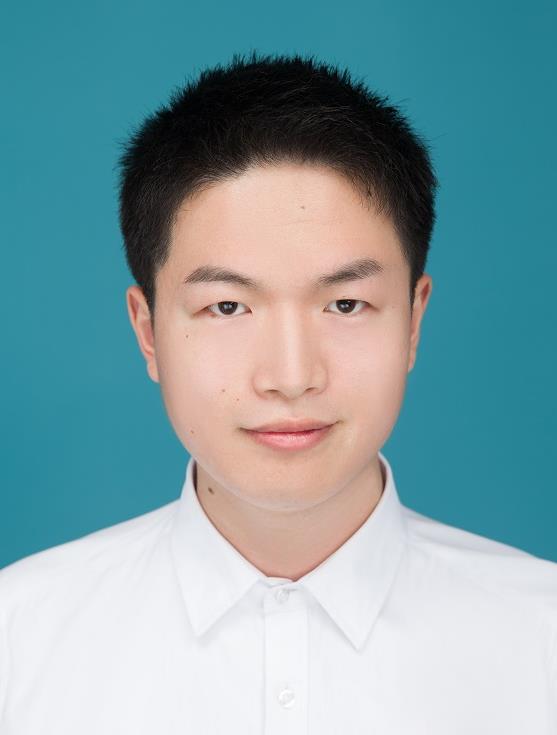}}]
	{Bingke Wang}	received the B.S. degree in control science and engineering from the University of Science and Technology of China (USTC), Hefei, China, in 2018, and is currently working toward the M.S. degree in control science and engineering at the USTC.
	
	His current research interests include lane detection and segmentation.
\end{IEEEbiography}







\end{document}